\documentclass[lettersize,journal]{IEEEtran}
\usepackage{amsmath,amsfonts}
\usepackage{algorithm}
\usepackage{array}
\usepackage{textcomp}
\usepackage{stfloats}
\usepackage{url}
\usepackage{verbatim}
\usepackage{graphicx}
\usepackage{cite}
\usepackage{algorithmicx}
\usepackage{algpseudocode}
\usepackage{multirow}
\usepackage{booktabs}
\usepackage{colortbl} 
\usepackage{xcolor}
\usepackage{subfigure} 
\usepackage{amssymb}

\def \A {{\mathbf{A}}}

 \def \D {\mathbf{D}}

 \def \H {\mathbf{H}} 
\def \I {\mathbf{I}}  
  \def \0 {\mathbf{0}}
 \def \Q {\mathbf{Q}}

\def \r {\mathbf{r}} \def \R {\mathbf{R}}
 \def \S {\mathbf{S}}
 \def \T {\mathbf{T}}

 \def \U {\mathbf{U}}
 \def \P {\mathbf{P}}

 \def \W {\mathbf{W}} 
  \def \I {\mathbf{I}}

\def \x {\mathbf{x}}
\def \X {{\mathbf{X}}}

\def \y {\mathbf{y}}
\def \Z {\mathbf{Z}} \def \z {\mathbf{z}}
 \def \Eps \mathbf{\varepsilon}

\def \diag {\mbox{diag}}

  \def \ce {{\mathcal E}}
  \def \cg {{\mathcal G}}

\makeatletter

\newcommand{\Rmnum}[1]{\expandafter\@slowromancap\romannumeral #1@}
\makeatother

\newcommand{\bmu}{\boldsymbol{\mu}}

\begin{document}

\title{Redundancy-Free Self-Supervised Relational Learning for Graph Clustering}

\author{Si-Yu~Yi,~Wei~Ju$^\dag$,~\IEEEmembership{Member,~IEEE,}~Yifang Qin,~Xiao Luo,~Luchen Liu,~Yong-Dao Zhou$^\dag$,~and~Ming~Zhang$^\dag$
\thanks{This paper is partially supported by the National Natural Science Foundation of China with Grant (NSFC Grant Numbers 62306014, 62106008, 62276002, 11871288 and 12131001), the China Postdoctoral Science Foundation with Grant No. 2023M730057, and the Fundamental Research Funds for the Central Universities, LPMC, and KLMDASR.}
\IEEEcompsocitemizethanks{\IEEEcompsocthanksitem Si-Yu Yi and Yong-Dao Zhou are with School of Statistics and Data Science, Nankai University, Tianjin, 300071, China.
(e-mail: siyuyi@mail.nankai.edu.cn, ydzhou@nankai.edu.cn)
\IEEEcompsocthanksitem Wei Ju, Yifang Qin, Luchen Liu and Ming Zhang are with School of Computer Science, Peking University, Beijing, 100871, China.
(e-mail: $\{$juwei,qinyifang,liuluchen,mzhang$\_$cs$\}$@pku.edu.cn)
\IEEEcompsocthanksitem Xiao Luo is with Department of Computer Science, University of California, Los Angeles, 90095, USA.
(e-mail: xiaoluo@cs.ucla.edu)
\IEEEcompsocthanksitem Corresponding authors: Wei Ju, Yong-Dao Zhou and Ming Zhang 
}
}



\maketitle

\begin{abstract}
Graph clustering, which learns the node representations for effective cluster assignments, is a fundamental yet challenging task in data analysis and has received considerable attention accompanied by graph neural networks in recent years. However, most existing methods overlook the inherent relational information among the non-independent and non-identically distributed nodes in a graph. Due to the lack of exploration of relational attributes, the semantic information of the graph-structured data fails to be fully exploited which leads to poor clustering performance. In this paper, we propose a novel self-supervised deep graph clustering method named Relational Redundancy-Free Graph Clustering (R$^2$FGC) to tackle the problem. It extracts the attribute- and structure-level relational information from both global and local views based on an autoencoder and a graph autoencoder. To obtain effective representations of the semantic information, we preserve the consistent relation among augmented nodes, whereas the redundant relation is further reduced for learning discriminative embeddings. In addition, a simple yet valid strategy is utilized to alleviate the over-smoothing issue. Extensive experiments are performed on widely used benchmark datasets to validate the superiority of our R$^2$FGC over state-of-the-art baselines. Our codes are available at https://github.com/yisiyu95/R2FGC.

\end{abstract}


\section{Introduction}
\IEEEPARstart{C}{lustering}, as one of the most classical and fundamental components in machine learning and data mining communities, has attracted significant attention. It serves as a critical preprocessing step in a variety of real-world applications such as community detection~\cite{liu2020deep}, anomaly detection~\cite{sheng2019multi}, domain adaptation~\cite{tang2020unsupervised}, and representation learning~\cite{xu2021self,ju2023unsupervised,luo2022clear}. The underlying idea of clustering is to assign the samples to different groups such that similar samples are pulled into the same cluster while dissimilar samples are pushed into different clusters. Hence, clustering intuitively reflects the characteristics of the whole dataset, which could provide a priori information for various downstream domains, including computer vision and natural language processing.

Among many challenges therein, how to effectively partition the whole dataset into different clusters remains a fundamental yet open challenge such that the intrinsic distribution information of the dataset can be well preserved. To achieve this goal, a large number of advanced approaches have been developed over the past decades~\cite{ng2001spectral,vidal2011subspace}. Traditional clustering methods such as subspace clustering~\cite{vidal2011subspace} and spectral clustering~\cite{ng2001spectral} aim at projecting the data samples into a low-dimensional space coupled with additional constraint information so that the samples in the latent space can be clearly separated. However, the two-stage training paradigm of the traditional methods is typically sub-optimal since the representation learning and clustering are dependent on each other that should be jointly optimized. Moreover, traditional algorithms have limited model capacity that unavoidably limits their applicability and potential. Recently, benefiting from the strong representation capability of deep learning, massive deep clustering algorithms are proposed to show great potential and advantages over traditional approaches~\cite{caron2018deep,mukherjee2019clustergan,li2021contrastive,zhong2021graph,liu2023self}. The core essence of deep clustering is to group the data samples into different clusters through deep neural networks in an end-to-end fashion. In this way, clustering and representation learning are jointly optimized to learn clustering-friendly representations without manual feature extraction. For example, CC~\cite{caron2018deep} jointly learned effective representations and cluster assignments by leveraging the power of instance- and cluster-level contrastive learning in an end-to-end manner.


With the prevalence of graph-structured data, Graph Neural Networks (GNNs) have been extensively studied and achieved remarkable progress for many promising graph-related tasks and applications~\cite{wu2020comprehensive,ji2021survey,ju2023comprehensive}. One fundamental problem therein is graph clustering, which divides nodes in a graph into different clusters. GNNs can be well utilized for enhancing graph clustering performance to learn effective cluster assignments~\cite{shi2019robust,bo2020structural,peng2021attention,zhao2021graph,tu2021deep,zhu2022collaborative,zhang2022embedding,he2022parallelly,liu2022deep,peng2023dual}. Recently, there has been an increasing body of approaches on graph clustering. For example, SDCN~\cite{bo2020structural} firstly incorporated the topological structure knowledge into deep clustering accompanied by autoencoder (AE)~\cite{hinton2006reducing} and GNN. To better combine node attributes and structure information, DFCN~\cite{tu2021deep} improved the graph autoencoder (GAE)~\cite{kipf2016variational} and developed a fusion mechanism to dynamically integrate both sides for robust target distribution generation. Based on AE, AGCC~\cite{he2022parallelly} incorporated the attention mechanism to fuse learned node representations and leveraged a self-supervised mechanism to guide the clustering optimization procedure.

Despite the promising achievements of previous methods, a vast majority of existing graph clustering approaches still suffer from two key limitations: (i) \textbf{Neglect the exploration of relational information.} Most existing GNN-based methods only use message passing to aggregate neighboring information of the nodes in a graph. The high-order attributive and structural relationships of the non-IID graph-structured data are not well exploited, which leads that the underlying distribution information cannot be well revealed for meaningful representations; (ii) \textbf{Fail to reduce redundant information.} Many  clustering methods mainly focus on exploring graph information from multiple perspectives, unavoidably incorporating much redundant information into the learned representations, while the redundancy reduction is not taken into account, which prevents obtaining discriminative representations and excellent clustering performance. As such, it is highly promising to develop an approach that can fully explore the intrinsic relational information among nodes and decrease the redundant information for effective cluster assignments.

Towards this end, this paper proposes a novel deep clustering method called Relational Redundancy-Free Graph Clustering (R$^2$FGC). The key idea of R$^2$FGC is to exploit attribute- and structure-level relational information among the nodes from both global and local views in a redundancy-free manner. To achieve the goal effectively, R$^2$FGC first learns compact representations from an AE and a GAE to explore the attributive and structural information from complementary perspectives. Then, the relational information is extracted based on the learned representations from global and local views. Moreover, to fully benefit from the extracted relations, we preserve the consistent relationship such that the relational information for the same node is invariant to augmentations, whereas the correlations of the relational distribution for different nodes are reduced for learning discriminative representations. Further, R$^2$FGC combines the redundancy-free relational learning from both attribute and structure levels with an augmentation-based fusion mechanism to optimize the embedded representations in a self-supervised fashion. Comprehensive experiments are conducted to show that the proposed method can greatly improve the clustering performance compared with the existing state-of-the-art approaches over multiple benchmark datasets. To summarize, the main contributions of our work are as follows:

\begin{itemize}
\item \textbf{General Aspects:} 
This paper studies the inherent relational learning for non-IID graph-structured data and explores redundancy-free representations based on relational information for the graph clustering task. 
\item \textbf{Novel Methodologies:} We propose a novel approach to exploit attribute- and structure-level relational information among the nodes, which aims to extract augmentation-invariant relationships for the same node and decrease the redundant correlations between different nodes. Our R$^2$FGC is beneficial to obtain effective and discriminative representations for clustering. 
\item \textbf{Multifaceted Experiments:} We perform extensive experiments on various commonly used datasets to demonstrate the effectiveness of the proposed approach. 
\end{itemize}


\section{Related Work}
\label{sec::related}


\subsection{Graph Neural Networks}
Recent years have witnessed great progress in Graph Neural Networks (GNNs) and achieved state-of-the-art performance. The concept of GNNs was proposed \cite{scarselli2008graph} before 2010 and has become an ever-increasing theme. A general paradigm of GNNs is to iteratively update node representations by aggregating neighboring information based on message-passing~\cite{gilmer2017neural}. Representative method Graph Convolutional Network (GCN)~\cite{kipf2016semi} extended the classical convolutional neural networks to the case of graph-structured data. Subsequent work Graph Attention Network (GAT)~\cite{velivckovic2017graph} further leveraged the attention mechanism~\cite{vaswani2017attention} to dynamically aggregate the features of neighbors. With the powerful capability of GNNs, the learned graph representations can be used to serve a variety of downstream tasks, such as node classification~\cite{kipf2016semi,yuan2023learning}, graph classification~\cite{ying2018hierarchical,ju2023tgnn,ju2022kgnn}, and graph clustering~\cite{bo2020structural,ju2022glcc}.

\subsection{Deep Clustering}
The goal of deep clustering focus on utilizing the excellent representation ability of deep learning to serve the clustering process, which has achieved remarkable progress. Existing methods can be categorized into three main groups based on the training objectives: (i) reconstruction based methods, (ii) self-augmentation based methods, and (iii) spectral clustering based methods. The first group uses the AE to reconstruct the original input, which incorporates desired constraints on feature embeddings in the latent space. For instance, DEC~\cite{xie2016unsupervised} iteratively conducted the process of representation learning and clustering assignments via minimizing the Kullback-Leibler divergence. To preserve important data structure, IDEC~\cite{guo2017improved} introduced AE to improve the clustering so that the local structure of data generating distribution can be maintained. The second group aims to encourage the consistency between original samples and their augmented samples by optimizing the networks. For example, IIC~\cite{ji2019invariant}  sought to achieve the consistency of assignment probabilities by maximizing the mutual information of paired samples. The third group aims at constructing a robust affinity matrix for effective data partitioning. For instance, RCFE~\cite{li2018rank} utilized the idea of rank constraints and clusters data points in a low-dimensional subspace. Li et al.~\cite{li2018dynamic} utilized multiple features to construct affinity graphs for spectral clustering.

Benefiting from the breakthroughs of GNNs on graph-structured data, GNNs are capable of organically integrating node attributes and graph structures in a united way, and have emerged as a promising way for graph clustering. The basic idea is to group the nodes in the graph into several disjoint clusters. Similar to the deep clustering methods, a vast majority of existing graph clustering approaches~\cite{wang2017mgae,wang2019attributed,zhang2022embedding,he2022parallelly,pan2019learning,fan2020one2multi,bo2020structural,zhu2022collaborative} also continue the paradigm of AE, in which the GAE and the variational GAE (VGAE) are leveraged to operate on graph-structured data. For example, to dynamically learn the importance of the neighboring
nodes to the center node, DAEGC~\cite{wang2019attributed} employed the GAE to capture a compact representation by encoding the graph structures and node attributes. EGAE~\cite{zhang2022embedding} learned the explainable representations based on the GAE that can be also used for various tasks. Compared with previous methods, our work further explores graph clustering by simultaneously preserving the relational similarity and reducing the redundancy of the learned representations based on both AE and GAE.

\subsection{Self-supervised Learning}
Recently, self-supervised learning (SSL) revitalizes and has achieved superior performance across numerous domains. This technique is completely free of the need for explicit labels~\cite{liu2021self}, due to its powerful capability in learning effective representations from unlabeled data. The core procedure of SSL is first designing a domain-specific pretext task and training the networks on the pretext task, such that the learned representations can be more discriminative and applicable. Recently, many SSL approaches have been proposed to marry the power of SSL and deep learning~\cite{chen2020simple,he2020momentum,chen2020improved,grill2020bootstrap}, and have shown competitive performance in various downstream application~\cite{yuan2020self,yan2021zeronas,ju2022kernel,lee2022relational,ju2023few}. For example, SimCLR~\cite{chen2020simple} employed multiple data augmentations and a learnable nonlinear transformation to train an encoder, such that the model can pull the feature representations from the same samples together. To alleviate the issue of the large batch size of SimCLR, MoCo~\cite{he2020momentum} introduced a moving-averaged encoder to set up a dynamic dictionary for SSL. Furthermore, our proposed R$^2$FGC inherits the advantages of SSL to preserve the consistent relation and reduce the redundant information among nodes from global and local views for graph clustering.

\section{Notations and problem definition}\label{sec::definition}

In this section, we first briefly give the basic notations and formal terminologies in a graph. Then we introduce the concept of Graph Convolutional Network (GCN) and the problem formalization of graph clustering. 

{\bf Notations.} Let $\cg=(V,E,\X)$ denote an arbitrary undirected graph, where $V=\{v_1,\ldots,v_n\}$ is the vertex set with $n$ nodes, $E$ is the edge set,  $\X=(\x_1,\ldots,\x_n)^{\top} \in \mathbb{R}^{n \times d}$ is the node attribute matrix with $\x_i$ corresponding to node $i$ for $i=1,\ldots,n$, and $d$ is the dimensionality of the node attributes. $\A = (a_{ij}) \in \mathbb{R}^{n \times n}$ denote the adjacency matrix which is generated according to the adjacency relationships in $E$, and $a_{ij}=1$ if $(v_i,v_j)\in E$, i.e., there is an edge from node $v_i$ to node $v_j$, otherwise $a_{ij}=0$. The adjacency matrix can be normalized by $\S=\tilde{\D}^{-1/2}\tilde{\A}\tilde{\D}^{-1/2}$, where $\tilde{\A}=(\tilde{a}_{ij})=\A+\I$, $\I \in \mathbb{R}^{n\times n}$ is the identify matrix for adding self-connections, and $\tilde{\D}=\diag(\tilde{d}_1,\ldots,\tilde{d}_n)$ is the corresponding degree matrix with $\tilde{d}_i = \sum_{j=1}^n \tilde{a}_{ij}$. 

{\bf Graph Convolutional Network.} GCN generalizes the classical Convolutional Neural Networks to the case of graph-structured data. It utilizes the graph directly and learns new representations by aggregating the information of a node and its neighbors. In general, a layer of GCN has the form
\begin{equation*}
\H^{(l+1)} = \sigma(\S\H^{(l)}\W^{(l)}), 
\end{equation*}
where $\H^{(0)}$ is the input data, $\sigma(\cdot)$ is an activation function, such as Tanh and ReLU, $\H^{(l)}$ and $\W^{(l)}$ are the learned embedded representation and the trainable weight matrix in the $l$-th ($l>0$) layer, respectively.

{\bf Graph Clustering.} Given an unlabeled graph with $n$ nodes, the target of the graph clustering task is to divide these unlabeled nodes into $K$ disjoint clusters $\{C_1,\ldots, C_K\}$ based on a well-learned embedding matrix $\tilde{\Z} \in \mathbb{R}^{n \times d'}$, where $d'$ is the number of dimension of the latent embeddings. The nodes in the same cluster are highly similar and cohesive, while the nodes in different clusters are discriminative and separable. 


\section{The proposed method}\label{sec::model}

\begin{figure*}
    \centering
    \includegraphics[width=1\textwidth]{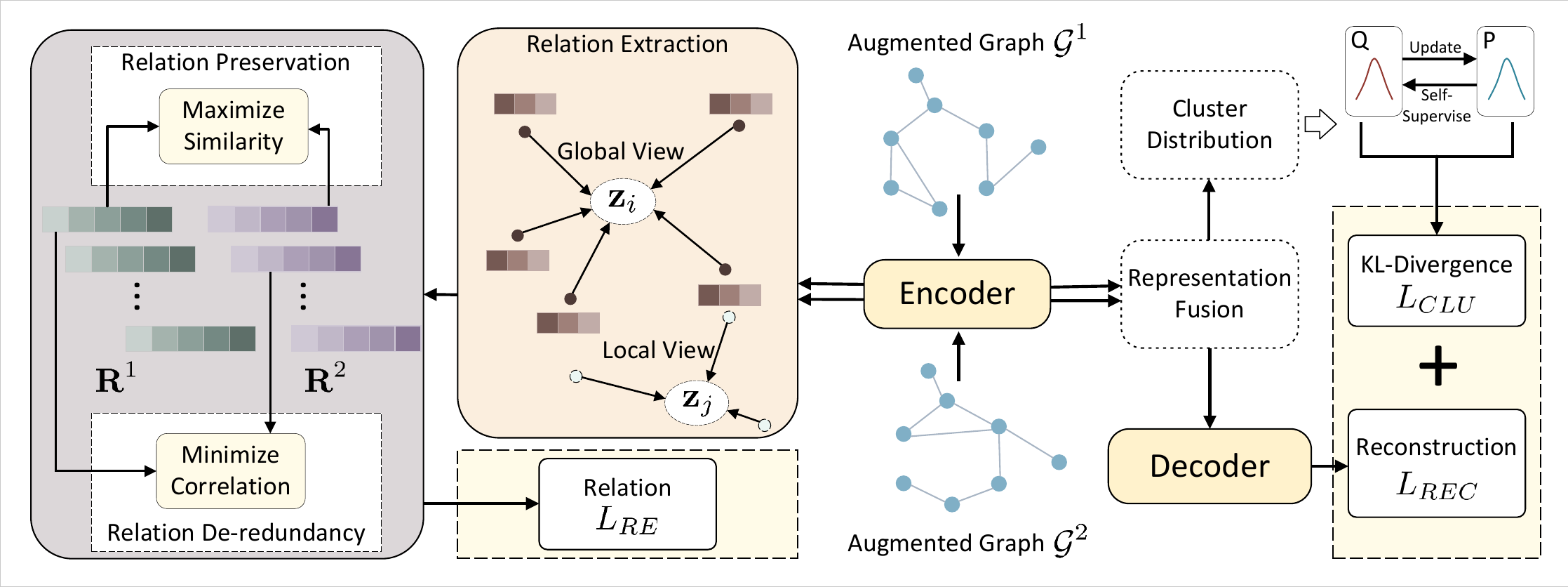}
    \caption{Framework overview of the proposed method R$^2$FGC. Relational learning and representation fusion are performed to jointly guide the self-supervised graph clustering based on the latent representations from the encoders of AE and GAE. The relation preservation and de-redundancy contribute to exploring inherent node relationship and filter redundancy relation to learn effective and discriminative representations.} 
    \label{fig:framework}
\end{figure*}

In this section, we introduce our proposed method named Relational Redundancy-Free Graph Clustering (R$^2$FGC). R$^2$FGC mainly contains four parts, i.e., attribute- and structure-level representation learning module, relation preservation and de-redundancy module, augmentation-based representation fusion module, and joint optimization module for graph clustering. Figure \ref{fig:framework} shows the framework overview of the proposed R$^2$FGC. In the following, we present the four components and the complexity analysis for R$^2$FGC.

\subsection{Attribute- and Structure-level Learning Module}\label{aegae}

AE can reasonably explore the node attribute information, whereas the GAE can effectively capture the topological structure information. To gain a more comprehensive embedding and a better performance on downstream tasks, we consider both AE and GAE to reconstruct the input and learn fusional representation. 

The AE module feeds the attribute information into the multi-layer perceptrons and extracts the latent representations by minimizing the reconstruction loss between the input raw data and the reconstructed data. The corresponding optimization objective is formalized as 
\begin{align}\label{ae}
\min~ L_{AE} &=  \frac{1}{n}\vert\vert \X - \hat{\X}_{AE} \vert\vert^2_{F},  \\
s.t. ~ \Z_{AE} &= \phi_{e}(\X), \nonumber\\
 \hat{\X}_{AE} &= \phi_{d}(\Z_{AE}), \nonumber
\end{align}
where $\X \in \mathbb{R}^{n\times d}$ is the input attribute matrix and $\hat{\X}_{AE} \in \mathbb{R}^{n\times d}$ is the reconstructed data, $\vert\vert\cdot\vert\vert_{F}$ is the Frobenius norm, $
\Z_{AE} \in \mathbb{R}^{n \times d'}$ is the learned latent representation in AE, $\phi_{e}$ and $\phi_{d}$ are the encoder and decoder networks, respectively. 

In the GAE module, following the improved version in \cite{tu2021deep}, a multi-layer GCN is adopted to reconstruct the adjacency matrix and the attribute information. The corresponding reconstruction loss is formalized as 
\begin{align}\label{gae}
\min~ L_{GAE} &=  \frac{\alpha}{n}\vert\vert \S-\hat{\S} \vert\vert^2_{F} + \frac{1}{n}\vert\vert \S\X-\hat{\X}_{GAE} \vert\vert^2_{F},  \\
s.t. ~ \H_{e}^{(l+1)} &= \sigma(\S \H_{e}^{(l)} \W_{e}^{(l)}), \nonumber\\
 \H_{d}^{(l+1)} &= \sigma(\S \H_{d}^{(l)} \W_{d}^{(l)}), \nonumber\\
 \H_{e}^{(0)} &= \X, \nonumber
\end{align}
where $\alpha$ is a pre-defined hyper-parameter, $\S$ is the normalized adjacency matrix, 
$\hat{\S}$ is the reconstructed adjacency matrix produced by fusing respective inner products of the learned latent representation $\Z_{GAE}\in \mathbb{R}^{n\times d'}$ resulting from the graph encoder and the attribute representations $\hat{\X}_{GAE}$ (i.e., the reconstructed weighted attribute matrix) resulting from the graph decoder, 
$\W_{e}^{(l)}$ and $\W_{d}^{(l)}$ are the layer-specific trainable weight matrices in the $l$-th graph encoder and decoder layers, respectively. 
The detailed fusion mechanism is discussed in the following Section \ref{fusion}, which unites the embedded representations from both AE and GAE to promote latent presentation learning in the graph augmentation fashion.

\subsection{Relation Preservation and De-redundancy Module}\label{relation}

In this module, we learn the inherent relational information among the nodes based on augmentations on a given graph. One of the basic ideas is to preserve the similarity of the relational information from two augmented views, while the latent representation of the same node can vary after graph augmentation. Hence, we aim to increase the consistency of the relational information in each positive pair. It allows fine-grained mining of the node relationship. On the other hand, it is necessary to improve the discriminative capability of the resulting representations for graph clustering, thus we also decrease the correlation of the relational information in each negative pair. In the following, we first introduce the adopted graph augmentation strategies and relation extraction methods. Then, we describe the details of the subsequent relation preservation and relation de-redundancy.

Based on the given graph, we first construct two different graph views through augmentations, including 
\begin{itemize}

\item {\bf Attribute perturbation.} For each value in the attribute matrix, we disturb it by multiplying a Gaussian random number with a small variance. This strategy performs a slight disturbance on the node features, which would not essentially change the semantic information. 

\item {\bf Edge deletion.} We remove some edges based on the node similarity obtained from the pre-learned latent embeddings. For each node, the edges that connect the nodes with low similarity are dropped in a certain proportion. Compared with random deletion, more semantic information can be preserved by referring to the node similarity. 

\item {\bf Graph diffusion.} We transform the adjacency matrix to a diffusion matrix by leveraging graph diffusion \cite{klicpera2019diffusion}, which contributes to providing additional local information. Technically, given the transition matrix $\T$, the graph diffusion matrix $\U$ is formulated as 
\begin{equation*}\label{diff_mat}
\U = \sum_{j=0}^{\infty} \theta_j \T^j,
\end{equation*}
where $\theta_j$ is the weight coefficient. 
We adopt the personalized PageRank \cite{page1999pagerank} to characterize graph diffusion, which is a special case. Specifically, $\T$ is chosen as the normalized adjacency matrix $\S$ and $\theta_j=\eta(1-\eta)^j$ with teleport probability $\eta \in (0,1)$. Then, the resulting diffusion matrix $\U$ has the form
\begin{equation}\label{ppr_mat}
\U = \eta(\I-(1-\eta)\S)^{-1}. 
\end{equation}

\end{itemize}

After obtaining two augmented graph views $\cg^1 = \{\X^{1}, \S^{1}\}$ and $\cg^2 = \{\X^{2},\S^{2}\}$, we perform AE and GAE on $\X^{1}$ and $\X^{2}$, which generates the attribute-level latent representations $\Z_{AE}^1, \Z_{AE}^2$ and the structure-level latent representations $\Z_{GAE}^1, \Z_{GAE}^2$. 
To meticulously characterize the relational information, we explore the similarities of each node to some anchor nodes from both global and local perspectives based on these representations. 

{\bf Extraction of Global Anchors.} For capturing the global relationship of a query node $v_i\in V$, the target is to sample diverse anchors from the whole graph nodes. Due to the neighborhood aggregation mechanism in GNNs, we argue that the high-degree nodes may receive more information when passing messages, while the low-degree nodes would receive less information. This may result in poor representations for the nodes with low degrees. Hence, we perform non-uniform sampling on the nodes to balance the qualities of the representations for low- and high-degree nodes. Specifically, we adopt an inverse degree-weighted distribution for sampling anchors, which puts a larger sampling probability on a lower-degree node. The sampling weight and probability for each node, respectively, are as follows, 
\begin{align*}
w_i & = \beta^{\log(\tilde{d}_{i}+1)}, \\
p_i & =  \frac{w_i}{\sum_{v_j \in V} w_j}, \text{for any~} v_i \in V,  
\end{align*}
where $\beta \in (0,1)$ is a hyper-parameter to control the skewness of the distribution, and $\tilde{d}_{i}$ is the degree of node $v_i$. Moreover, quasi-Monte Carlo (QMC) sampling methods usually can achieve a higher convergence rate than Monte Carlo (MC) methods \cite{L09}. Hence, based on the defined discrete distribution, we perform multinomial sampling in the QMC fashion \cite{yi2023global,yi2023model}.  Instead of the uniform random number (the MC fashion), we leverage the randomized one-dimensional low-discrepancy point set $\{(2i-1)/(2M_1)+\omega \mod 1 \in [0,1]: \omega \sim U(0,1), i=1,\ldots, M_1\}$ to do multinomial sampling on the discrete distribution in each training epoch. Randomization is used to avoid the same sample in different epochs and increase the randomness for extracting more diverse anchors. 
For each node $v_i \in V$, we denote the index set of the sampled anchors from the global view as $A_i^{g}$ and $\vert A_i^{g}\vert = M_1$. 

{\bf Extraction of Local Anchors.} To fully explore the relational information, besides the global anchor sampling, we also concentrate on the local relational information. Graph diffusion removes the restriction of using only the direct neighbors and alleviates the problem of noisy and often arbitrarily defined edges. 
It leads that the diffusion matrix $\U$ in \eqref{ppr_mat} can acquire richer structural information in the local view compared with traditional GNNs. Hence, we leverage graph diffusion to generate the local anchors according to the scores in $\U$. Specifically, the values in the $i$-th row of $\U$ can reflect the influence between node $v_i$ and all the other nodes. We select the nodes with the $M_2$ largest scores in the $i$-th row of $\U$ as the local anchors of node $v_i$. It makes that the local anchors of $v_i$ share similar semantic information to $v_i$, which allows us to extract more effective local relations. For each node $v_i\in V$, we denote the index set of the local anchors as $A_i^{l}$ and $\vert A_i^{l} \vert = M_2$. 

Based on these global- and local-view anchor sets $A_i^{g}, A_i^{l}, i=1,\ldots,n$, we extract the relational information of the nodes in the sense of similarity. 
We use the AE latent representations 
$\Z_{AE}^1=(\z^1_{AE,1},\ldots,\z^1_{AE,n})^{\top}, 
\Z_{AE}^2=(\z^2_{AE,1},\ldots,\z^2_{AE,n})^{\top}$ to illustrate the detailed process. 
Specifically, given a query node $v_i \in V$, we calculate the similarities between the embedded representation of $v_i$ in $\Z_{AE}^1$ and the embeddings of these anchors in $\Z_{AE}^2$ by
\begin{align*}
r^{1}_{g}(i,k_g) & =(\z^1_{AE,i})^{\top}\z^2_{AE,k_g}, k_g \in A_i^{g}, \\
r^{1}_{l}(i,k_l) & =(\z^1_{AE,i})^{\top}\z^2_{AE,k_l}, k_l \in A_i^{l}.  
\end{align*}
Similarly, we also calculate the similarities between the embedding of $v_i$ in $\Z_{AE}^2$ and those of the anchors in $\Z_{AE}^2$ by 
\begin{align*}
r^{2}_{g}(i,k_g) & =(\z^2_{AE,i})^{\top}\z^2_{AE,k_g}, k_g \in A_i^{g}, \\
r^{2}_{l}(i,k_l) & =(\z^2_{AE,i})^{\top}\z^2_{AE,k_l}, k_l \in A_i^{l}.  
\end{align*}
Hereafter, let $\r^u_{c}(i)$ be the relation vector composed by $r^{u}_{c}(i,k)$ with $k$ traversing the whole index set $A_i^{c}$ of node $v_i, i \in\{1,\ldots,n\}$, and $\R^u_c = (\r^u_{c}(1), \ldots, \r^u_{c}(n))^{\top}$ be the relation matrix for any $u\in\{1,2\}, c\in\{g,l\}$. 

{\bf Relation Preservation.} To make the relational information invariant to augmentation, we maximize the proximity of $\r^1_{c}(i)$ and $\r^2_{c}(i)$ from both global and local views, i.e., we maximize the attribute-level relational similarities of all the positive pairs under augmentation, 
which are formulated by 
\begin{equation*}\label{re_glo}
R_{AE}^{g}= \frac{1}{n}\sum_{i=1}^n \left( \frac{\r^1_{g}(i)^{\top} \r^2_{g}(i)}{\vert\vert \r^1_{g}(i)\vert\vert\cdot\vert\vert \r^2_{g}(i)\vert\vert} \right)^2
\end{equation*}
and
\begin{equation*}\label{re_loc}
R_{AE}^{l}= \frac{1}{n} \sum_{i=1}^n \left( \frac{\r^1_{l}(i)^{\top} \r^2_{l}(i)}{\vert\vert \r^1_{l}(i)\vert\vert\cdot\vert\vert \r^2_{l}(i)\vert\vert} \right)^2. 
\end{equation*}
We can similarly obtain the structure-level relational similarities $R_{GAE}^{g}$ and $R_{GAE}^{l}$ 
corresponding to 
GAE from both views. This operation helps to learn representations that are more reflective of the relationships between the attribute and topological information of all the nodes.

{\bf Relation De-redundancy.} In addition, besides preserving the relational similarity under augmentations, the discriminative capability of the latent representation is also important for the downstream graph clustering task. Hence, we decrease the correlations of the relation vectors for different nodes from both global and local views. It contributes to filtering redundant information and improving the separating capability for better clustering performance. Specifically, we minimize the attribute-level relational correlations of all the negative pairs,  
which are formulated as follows, 
\begin{equation*}\label{loss_glo}
C_{AE}^{g}=\frac{1}{n(n-1)}\sum_{i,j=1,i\neq j}^n\left( \frac{\r^1_{g}(i)^{\top} \r^2_{g}(j)}{\vert\vert \r^1_{g}(i)\vert\vert\cdot\vert\vert \r^2_{g}(j)\vert\vert} \right)^2
\end{equation*}
and
\begin{equation*}\label{loss_loc}
C_{AE}^{l}=\frac{1}{n(n-1)}\sum_{i,j=1,i\neq j}^n\left( \frac{\r^1_{l}(i)^{\top} \r^2_{l}(j)}{\vert\vert \r^1_{l}(i)\vert\vert\cdot\vert\vert \r^2_{l}(j)\vert\vert} \right)^2. 
\end{equation*}
In like manner, we can obtain the corresponding structure-level loss under GAE from global and local views, denoted by $C_{GAE}^{g}$ and $C_{GAE}^{l}$, respectively.  

Based on the above discussion, we can capture the augmentation-invariant relational information and conduct redundancy-free relational learning by minimizing the total relation loss $L_{RE}=L_{REA}+L_{REG}$ with 
\begin{align}\label{loss_re}
L_{REA}  = & C_{AE}^{g}+C_{AE}^{l} \nonumber \\
& -R_{AE}^{g}-R_{AE}^{l}, \\
L_{REG} = & C_{GAE}^{g}+C_{GAE}^{l} \nonumber \\
& -R_{GAE}^{g}-R_{GAE}^{l}. \nonumber 
\end{align}

The loss $L_{RE}$ takes into account both efficient representation learning and reduction of redundant information upon the relation extraction of the nodes, which allows for better guidance of downstream tasks.

\subsection{Augmentation-based Representation Fusion Module}\label{fusion}

In this section, to obtain fine-grained representations of the nodes, we discuss the fusion mechanism of the attribute- and structure-level latent representations based on augmentations. 
First, we take a weighted summation of the four parts to fuse the  embedded representations from the two levels as follows, 
\begin{equation*}
\tilde{\Z}_{c}=\W_1\odot(\Z_{AE}^1+\Z_{AE}^2)+\W_2\odot(\Z_{GAE}^1+\Z_{GAE}^2), 
\end{equation*}
where $\W_1,\W_2 \in \mathbb{R}^{n\times d'}$ are trainable weight matrices to control the importance of the two types of representations and $\odot$ is the Hadamard product. 
Based on $\tilde{\Z}_{c}$, we further blend the embeddings from both global and local views to refine the fused information. From the local view, we adopt the neighborhood aggregation operation on $\tilde{\Z}_{c}$ to enhance the local information, whereas, from the global view, we utilize the self-correlation matrix of the nodes characterized by $\tilde{\Z}_{c}$ to improve the exploitation of the global information, which is normalized by the softmax function.  Specifically, the final formula of the fused representation is 
\begin{equation}\label{rep}
\tilde{\Z}=\delta \S \tilde{\Z}_c + softmax(\S\tilde{\Z}_c \tilde{\Z}_c^{\top}\S^{\top})\S \tilde{\Z}_c, 
\end{equation}
where $\delta$ is a trainable weight parameter. 
With $\tilde{\Z}$, we can obtain the reconstructed attribute matrix $\hat{\X}_{AE}$ in \eqref{ae} and weighted attribute matrix $\hat{\X}_{GAE}$ in \eqref{gae} by feeding $\tilde{\Z}$ into the decoders of AE and GAE, respectively. The reconstructed adjacency matrix is calculated by fusing the self-correlations of the learned representations in GAE, which is formulated as 
\begin{equation*}
\hat{\S}=\frac{1}{2}(\Z_{GAE}^1(\Z_{GAE}^1)^{\top}+\Z_{GAE}^2(\Z_{GAE}^2)^{\top})+\hat{\X}_{GAE} \hat{\X}_{GAE}^{\top}. 
\end{equation*}
The above fusion process is similar to \cite{tu2021deep}.

In addition, under the neighbor aggregation mechanism, GCN updates node representations by aggregating information from the neighbors. However, when stacking multiple layers, the learned representations would become indistinguishable, seriously degrading the performance, which is the so-called over-smoothing issue \cite{oono2019graph,chen2020measuring}. 
Hence, it is important to balance the message aggregation ability and over-smoothing issue.  
To alleviate the problem in GAE, we incorporate a novel propagation-regularization loss to enhance  information capturing while alleviating over-smoothing defined as
\begin{equation*}\label{p_reg}
L_{PR}=\sum_{\H \in \ce} \nu(\H, \S\H), 
\end{equation*}
where $\ce$ contains the embedding matrix in each layer of both the encoder and decoder in GAE and $\nu(\cdot)$ is the metric function, such as the cross entropy, Kullback-Leibler (KL) divergence, and the Jensen-Shannon divergence. Propagation regularization simulates a deep GCN by supervision at a low cost, which enables current embeddings to capture further information contained in the deeper layer. 
Compared with directly increasing the GCN layers, we can more finely balance the information capture ability and the over-smoothing problem by adjusting the weight of the loss. 

Thereby, the total reconstruction loss is computed by 
\begin{equation}\label{loss_rec}
L_{REC}=L_{AE}+L_{GAE}+\epsilon L_{PR},
\end{equation}
where $\epsilon$ is the pre-defined hyper-parameter to adjust the influence ratio, and the reconstruction losses $L_{AE}$ and $L_{GAE}$ in AE and GAE are defined in \eqref{ae} and \eqref{gae}, respectively.

\subsection{Joint Optimization Module for Graph Clustering}\label{cluster}

Graph clustering is essentially an unsupervised task with no feedback available as reliable guidance. To this end, we perform a clustering layer on the fused representation $\tilde{\Z}$ in \eqref{rep} and use the soft labels derived by a probability distribution as a self-supervised signal to jointly optimize the redundancy-free relational learning framework for graph clustering. 

First, by using the student's $t$-distribution as a kernel, we calculate the soft cluster assignment probabilities $\Q_1=(q_{1,ij}), \Q_2=(q_{2,ij}), \Q_3=(q_{3,ij})\in\mathbb{R}^{n \times K}$ upon the latent embeddings $\tilde{\Z}, (\Z_{AE}^1+\Z_{AE}^2)/2, (\Z_{GAE}^1+\Z_{GAE}^2)/2$, respectively, to measure the similarities between the latent representations and cluster centroids, i.e., each value indicates the probability of assigning the $i$th node to the $j$th cluster. For example, $q_{1,ij}$ is computed as follows, 
\begin{equation*}\label{q1ij}
q_{1,ij}=\frac{(1+\vert\vert \tilde{\z}_i-\bmu_j \vert\vert^2)^{-1}}{\sum_{k=1}^K (1+\vert\vert \tilde{\z}_i-\bmu_k \vert\vert^2)^{-1}}, 
\end{equation*}
where $\tilde{\Z}=(\tilde{\z}_1^{\top},\ldots,\tilde{\z}_n^{\top})^{\top}$ and $\bmu_j$'s are the cluster centroids. The $q_{2,ij}$ and $q_{3,ij}$ can be calculated similarly. The $\bmu_j$'s are initialized by performing $k$-Means on the pre-trained fused representation. When the network is well trained, we adopt the fusion-based assignment matrix $\Q_1$ to measure the cluster assignment probability of all the nodes, i.e., 
\begin{align}\label{res}
   y_i & = \text{argmax}_{j\in \{1,\ldots,K\}} q_{1,ij}, 
 \end{align}
 where $y_i$ is the predicted cluster of node $v_i$ for $i=1,\ldots,n$. 

\begin{algorithm}[t]
	\caption{Relational Redundancy-Free Graph Clustering}
	\label{code1}
	\begin{algorithmic}[1]
	\Require Attribute matrix $\X$; Adjacency matrix $\A$; Cluster number $K$; hyper-parameters $M_1,M_2$; 
	Maximum Iterations $I_{max}$;
	\Ensure Clustering result $\y$;
	\State Initialize the parameters in AE, GAE, the fusion part, and the cluster centroids; 
	\For{$i=1$ to $I_{max}$}
	\State Obtain $\{\X^{1}, \S^{1}\}$ and $\{\X^{2},\S^{2}\}$ by augmentation; 
	\State Update $\Z_{AE}^1, \Z_{AE}^2$ and $\Z_{GAE}^1, \Z_{GAE}^2$ by encoding $\X^{1}$ and $\X^{2}$ in AE and GAE; 
	\State Calculate $\R^1_g, \R^2_g, \R^1_l, \R^2_l$ based on AE and GAE in Section \ref{relation}; 
	\State Update $\tilde{\Z}_c$, $\tilde{\Z}$, $\hat{\S}$ in Section \ref{fusion} and obtain $\hat{\X}_{AE}$, $\hat{\X}_{GAE}$ in Section \ref{aegae}; 
	\State Calculate $\Q_1, \Q_2, \Q_3$, and $\P$ in Section \ref{cluster}; 
	\State Calculate the losses $L_{RE},L_{REC},L_{CLU}$ in \eqref{loss_re}, \eqref{loss_rec}, \eqref{loss_clu}, respectively;
	\State Conduct the backpropagation and update the whole network in the proposed R$^2$FGC by minimizing \eqref{total_loss};
	\EndFor
	\State Obtain the clustering result $\y$ with the fused representation $\tilde{\Z}$ by \eqref{res}; 
	\State \Return{$\y$};
	\end{algorithmic}
\end{algorithm}

Next, we introduce an auxiliary confident probability distribution $\P=(p_{ij})\in \mathbb{R}^{n\times K}$ to improve the confidence of the soft assignment, which is derived from $\Q_1$ and formulated as  
\begin{equation*}
p_{ij}=\frac{q_{1,ij}^2/\sum_{i=1}^n q_{1,ij}}{\sum_{k=1}^K (q_{1,ik}^2/\sum_{i=1}^n q_{1,ik})}. 
\end{equation*}
To make the data representation close to cluster centroids and improve cluster cohesion, we minimize the KL divergence loss between $\P$ and $\Q_1,\Q_2,\Q_3$ as follows, 
\begin{equation}\label{loss_clu}
L_{CLU}=\sum_{i=1}^n\sum_{j=1}^K p_{ij}\log \frac{p_{ij}}{(q_{1,ij}+q_{2,ij}+q_{3,ij})/3}.  
\end{equation}
By utilizing the confident distribution $\P$, the process self-supervises the cluster assignment without any label guidance. We integrate the latent representations from AE, GAE, and the fusion mechanism in the self-supervised clustering procedure to obtain more accurate clustering results. 

To sum up, the total loss $L$ in the whole framework of R$^2$FGC is composed of the relation loss, the reconstruction loss, and the self-supervised clustering loss, i.e.,
\begin{equation}
\label{total_loss}
L=L_{RE}+L_{REC}+\kappa L_{CLU}, 
\end{equation}
where $\kappa$ is a pre-defined hyper-parameter to balance the weight of the clustering loss. The training process of our proposed R$^2$FGC is summarized in Algorithm \ref{code1}.

\subsection{Computational Complexity Analysis}\label{compl}

For the scalability of large-scale datasets, we adopt the mini-batch stochastic gradient descent to optimize our method. Assume that the batch size is $B$ and the dimensions of each layer of AE and GAE are $\bar{d}_1,\ldots,\bar{d}_{L_1}$ and $\tilde{d}_1,\ldots,\tilde{d}_{L_2}$, respectively. Given a graph with $n$ nodes and $|E|$ edges, the dimension of the original attributes is $d$. The time complexities of AE and GAE are $O(n\sum_{i=1}^{L_1}\bar{d}_i\bar{d}_{i-1})$ and $O(|E|\sum_{i=1}^{L_2}\tilde{d}_i\tilde{d}_{i-1})$ with $\bar{d}_0=\tilde{d}_0=d$, respectively. For each batch, the complexity of the relation learning module is $O(B(B+d')(M_1+M_2))$ based on $d'$-dimensional latent representations. Moreover, we perform the representation fusion and propagation regularization in $O(B^2d'+B\log B)$ time and conduct the self-supervised clustering in $O(BK+B\log B)$ time with $K$ classes in the task. Hence, the total computational complexity of our method R$^2$FGC is 
$O(n\sum_{i=1}^{L_1}\bar{d}_i\bar{d}_{i-1}+
|E|\sum_{i=1}^{L_2}\tilde{d}_i\tilde{d}_{i-1}+
n(B+d')(M_1+M_2)+
n(Bd'+K))$, which is linearly related to the numbers of nodes and edges.

\section{Experiments}\label{sec::experiment}

In this section, we first introduce the experimental settings and 
then conduct experiments to validate the effectiveness of R$^2$FGC. We aim to answer the following research questions. 

\begin{itemize}
  \item {\bf RQ1:} Compared with state-of-the-art methods, does our method R$^2$FGC achieve better performance for self-supervised graph clustering?
  \item {\bf RQ2:} How do different components of the proposed method contribute to the clustering performance?
  \item {\bf RQ3:} How do the hyper-parameters in R$^2$FGC affect the final clustering performance?
   \item {\bf RQ4:} How is the convergence of the proposed model under different datasets?
  \item {\bf RQ5:} Is there any supplementary analysis that can illustrate the superiority of R$^2$FGC?
\end{itemize}

\subsection{Experimental Settings}\label{es}

{\bf Datasets.} For comparison, we perform the proposed method R$^2$FGC on five commonly used benchmark datasets. Four of them are graph datasets, including a paper network ACM\footnote{http://dl.acm.org/}, a shopping network AMAP\footnote{https://github.com/shchur/gnn- benchmark/raw/master/data/npz/\\amazon\_electronics photo.npz}, a citation network CITE\footnote{http://citeseerx.ist.psu.edu/index}, and an author network DBLP\footnote{https://dblp.uni-trier.de}; another is a non-graph dataset, i.e., a record dataset HHAR\cite{stisen2015smart}. Following \cite{bo2020structural}, for the non-graph data, the adjacency matrix is generated by the undirected $k$-nearest neighbor graph. Table \ref{t2} briefly summarizes the information of these benchmark datasets. 

\begin{table}[!t]
\caption{Description of the benchmark datasets}\label{t2}
\centering
\tabcolsep=10pt
\begin{tabular}{c|c|c|c|c}
\toprule
\toprule
Dataset & Type & Samples & Dimension & Classes \\
\midrule 
ACM & Graph & 3025 & 1870 & 3 \\
\midrule 
AMAP & Graph & 7650 & 745 & 8 \\
\midrule 
CITE & Graph & 3327 & 3703 & 6 \\
\midrule 
DBLP & Graph & 4057 & 334 & 4 \\
\midrule 
HHAR & Record & 10299 & 561 & 6 \\
\bottomrule 
\bottomrule 
\end{tabular}
\end{table}

{\bf Compared Methods.} To illustrate the superiority of our proposed R$^2$FGC, we compare its clustering performance with some state-of-the-art clustering methods, which are divided into four categories, i.e., the classical shallow clustering method $k$-means, the AE-based methods, the GCN-based methods, and the combination of AE and GCN. The AE-based methods contains AE\cite{hinton2006reducing}, DEC\cite{xie2016unsupervised}, and IDEC\cite{guo2017improved}. They convert the raw data to low-dimensional codes to learn feature representations by AE and then perform clustering over the learned latent embeddings. The GCN-based methods include GAE, VGAE\cite{kipf2016variational}, DAEGC\cite{wang2019attributed}, and  ARGA\cite{pan2019learning}. They adopt the GCN encoder to learn the node content and topological information for clustering. In addition, some methods combine AE and GCN to boost the embedded representations for clustering, which contains SDCN~\cite{bo2020structural}, DFCN\cite{tu2021deep} and AGCC~\cite{he2022parallelly}. These methods integrate GCN with AE from different perspectives to jointly train the clustering network. 

{\bf Training Procedure.} The training of our method R$^2$FGC includes two phases. First, following \cite{tu2021deep}, the AE and GAE are pre-trained independently for 30 epochs to minimize their respective reconstruction loss functions. Both sub-networks are integrated into a united framework for another 100 epochs to obtain the initial representations and cluster centroids. Then, we train the whole network for at least 300 epochs until convergence to minimize the total loss in \eqref{total_loss}. Following the compared methods, to alleviate the adverse influence of the randomness, we repeat the experiment 10 times to evaluate our method and report the mean values and the standard deviations (i.e., mean±std) of the considered metric values. We implement our method using PyTorch 1.8.0 and Pytorch Geometric 1.7.2, which can easily train GNNs for a variety of applications associated with graph-structured data. 

\begin{table*}[!t]
\caption{The clustering performance on five benchmark datasets (mean±std). The best results in all the methods and all the baselines are highlighted with \textbf{bold} and \underline{underline}, respectively.}\label{t3}
\centering
\tabcolsep=2.5pt
\renewcommand\arraystretch{1.3}
{\scriptsize
\begin{tabular}{c|c|ccccccccccc|>{\columncolor{gray!25}}c}
\toprule
\toprule
\textbf{Dataset}                    & \textbf{Metric} & \textbf{$\boldsymbol{k}$-Means} & \textbf{AE} & \textbf{DEC} & \textbf{IDEC} & \textbf{GAE} & \textbf{VGAE} & \textbf{DAEGC}                    & \textbf{ARGA}    & 
\textbf{SDCN} & \textbf{DFCN}  & \textbf{AGCC}  & \textbf{R$^2$FGC (Ours)} \\ 
\midrule
                                    & ACC             & 67.31±0.71       & 81.83±0.08  & 84.33±0.76   & 85.12±0.52    & 84.52±1.44   & 84.13±0.22    & 86.94±2.83     & 86.29±0.36   
& 90.45±0.18           & \underline{90.90±0.20} & 90.38±0.38 & {\bf 92.43±0.18}    \\
                                    & NMI             & 32.44±0.46       & 49.30±0.16  & 54.54±1.51   & 56.61±1.16    & 55.38±1.92   & 53.20±0.52    & 56.18±4.15    & 56.21±0.82    
& 68.31±0.25         & \underline{69.40±0.40} & 68.34±0.89 & {\bf 72.42±0.53}   \\
                                    & ARI             & 30.60±0.69        & 54.64±0.16  & 60.64±1.87   & 62.16±1.50    & 59.46±3.10   & 57.72±0.67    & 59.35±3.89    & 63.37±0.86    
& 73.91±0.40       & \underline{74.90±0.40} & 73.73±0.90 & {\bf 78.72±0.47}    \\
\multirow{-4}{*}{\textbf{ACM}}      & F1              & 67.57±0.74       & 82.01±0.08  & 84.51±0.74   & 85.11±0.48    & 84.65±1.33   & 84.17±0.23    & 87.07±2.79   & 86.31±0.35    
& 90.42±0.19       & \underline{90.80±0.20} & 90.39±0.39 & {\bf 92.45±0.18}    \\ 
\midrule
                                    & ACC             & 27.22±0.76       & 48.25±0.08  & 47.22±0.08   & 47.62±0.08    & 71.57±2.48   & 74.26±3.63    & 76.44±0.01     & 69.28±2.30    
& 53.44±0.81       & \underline{76.88±0.80} & 75.25±1.21 & {\bf 81.28±0.05}    \\
                                    & NMI             & 13.23±1.33       & 38.76±0.30  & 37.35±0.05   & 37.83±0.08    & 62.13±2.79   & 66.01±3.40    & 65.57±0.03   & 58.36±2.76    
& 44.85±0.83       & \underline{69.21±1.00} & 68.37±1.39 & {\bf 73.88±0.17}    \\
                                    & ARI             & 5.50±0.44        & 20.80±0.47  & 18.59±0.04   & 19.24±0.07    & 48.82±4.57   & 56.24±4.66    & \underline{59.39±0.02} & 44.18±4.41    
& 31.21±1.23            & 58.98±0.84     & 58.32±2.38   & {\bf 66.25±0.36}    \\
\multirow{-4}{*}{\textbf{AMAP}}     & F1              & 23.96±0.51       & 47.87±0.20  & 46.71±0.12   & 47.20±0.11    & 68.08±1.76   & 70.38±2.98    & 69.97±0.02    & 64.30±1.95    
& 50.66±1.49                         & \underline{71.58±0.31} & 70.04±1.63  & {\bf 75.29±0.32}    \\ 
\midrule
                                    & ACC             & 39.32±3.17       & 57.08±0.13  & 55.89±0.20   & 60.49±1.42    & 61.35±0.80   & 60.97±0.36    & 64.54±1.39                        & 61.07±0.49    
& 65.96±0.31       & \underline{69.50±0.20} & 68.08±1.44  & {\bf 70.60±0.45}    \\
                                    & NMI             & 16.94±3.22       & 27.64±0.08  & 28.34±0.30   & 27.17±2.40    & 34.63±0.65   & 32.69±0.27    & 36.41±0.86                        & 34.40±0.71    
& 38.71±0.32      & \underline{43.90±0.20} & 40.86±1.45 & {\bf 45.39±0.37}    \\
                                    & ARI             & 13.43±3.02       & 29.31±0.14  & 28.12±0.36   & 25.70±2.65    & 33.55±1.18   & 33.13±0.53    & 37.78±1.24                        & 34.32±0.70    
& 40.17±0.43      & \underline{45.50±0.37} & 41.82±2.03 & {\bf 47.07±0.30}    \\
\multirow{-4}{*}{\textbf{CITE}}     & F1              & 36.08±3.53       & 53.80±0.11  & 52.62±0.17   & 61.62±1.39    & 57.36±0.82   & 57.70±0.49    & 62.20±1.32                        & 58.23±0.31    
& 63.62±0.24        & \underline{64.30±0.20} & 60.47±1.57 & {\bf 65.28±0.12}    \\
\midrule
                                    & ACC             & 38.65±0.65       & 51.43±0.35  & 58.16±0.56   & 60.31±0.62    & 61.21±1.22   & 58.59±0.06    & 62.05±0.48     & 64.83±0.59    
& 68.05±1.81       & \underline{76.00±0.80} & 73.45±2.16 & {\bf 80.95±0.20}    \\
                                    & NMI             & 11.45±0.38       & 25.40±0.16  & 29.51±0.28   & 31.17±0.50    & 30.80±0.91   & 26.92±0.06    & 32.49±0.45  & 29.42±0.92    
& 39.50±1.34       & \underline{43.70±1.00} & 40.36±2.81  & {\bf 50.82±0.32}    \\
                                    & ARI             & 6.97±0.39        & 12.21±0.43  & 23.92±0.39   & 25.37±0.60    & 22.02±1.40   & 17.92±0.07    & 21.03±0.52    & 27.99±0.91   
& 39.15±2.01                              & \underline{47.00±1.50} & 44.40±3.79  & {\bf 56.34±0.42}    \\
\multirow{-4}{*}{\textbf{DBLP}}     & F1              & 31.92±0.27       & 52.53±0.36  & 59.38±0.51   & 61.33±0.56    & 61.41±2.23   & 58.69±0.07    & 61.75±0.67   & 64.97±0.66    
& 67.71±1.51       & \underline{75.70±0.80} & 71.84±2.02  & {\bf 80.54±0.19}    \\ 
\midrule
                                    & ACC             & 59.98±0.02       & 68.69±0.31  & 69.39±0.25   & 71.05±0.36    & 62.33±1.01   & 71.30±0.36    & 76.51±2.19   & 63.30±0.80   
& 84.26±0.17     & \underline{87.10±0.10} & 86.54±1.79  & {\bf 88.91±0.05}    \\
                                    & NMI             & 58.86±0.01       & 71.42±0.97  & 72.91±0.39   & 74.19±0.39    & 55.06±1.39   & 62.95±0.36    & 69.10±2.28   & 57.10±1.40    
& 79.90±0.09        & 82.20±0.10  &  \underline{82.21±1.78}      & {\bf 83.39±0.07}    \\
                                    & ARI             & 46.09±0.02       & 60.36±0.88  & 61.25±0.51   & 62.83±0.45    & 42.63±1.63   & 51.47±0.73    & 60.38±2.15   & 44.70±1.00  
& 72.84±0.09     & \underline{76.40±0.10} &  75.58±1.85  & {\bf 78.52±0.08}    \\
\multirow{-4}{*}{\textbf{HHAR}}   & F1              & 58.33±0.03       & 66.36±0.34  & 67.29±0.29   & 68.63±0.33    & 62.64±0.97   & 71.55±0.29    & 76.89±2.18  & 61.10±0.90    
& 82.58±0.08            & \underline{87.30±0.10}  & 85.79±2.48   & {\bf 89.23±0.06}    \\ 
\bottomrule
\bottomrule
\end{tabular}
}
\end{table*}

{\bf Parameter Settings.} For a fair comparison, we adopt the same parameter setting for AE and GAE as \cite{tu2021deep}, i.e., the layers of the encoder (/decoder) for AE and GAE are set to 4 and 3, respectively; the dimensions of the encoder (/decoder) for AE are set to 128, 256, 512, 20 in turn; the dimensions of the encoder (/decoder) for GAE are set to 128, 256, 20 in turn. The network is trained with the Adam optimizer. The learning rate is set to 5e-5 for ACM, 1e-4 for DBLP, and 1e-3 for AMAP, CITE, and HHAR. The hyper-parameters $M_1$ and $M_2$ are set to $\{$256, 8$\}$. 
Moreover, the parameters $\alpha,\eta,\beta,\epsilon,\kappa$ are set to 0.1, 0.2, 0.8, 5e3, 10, respectively. The optimization stops when the validation loss comes to a plateau.

{\bf Evaluation Metrics.} To evaluate the clustering performance of each compared method, we adopt four widely used evaluation metrics following \cite{bo2020structural}, i.e., Accuracy (ACC), Normalized Mutual Information (NMI), Average Rand Index (ARI), and Macro F1-score (F1). For each metric, a larger value implies a better clustering result. 

\subsection{Performance Comparison (RQ1)}\label{RQ1}

The experimental results of our method and eleven compared methods on five benchmark datasets are reported in Table \ref{t3}, in which the bold and underlined values indicate the best results in all the methods and all the baselines, respectively. From these results, we have the following observations.

\begin{table*}[!t]
\caption{The ablation study on five benchmark datasets (mean±std). The results show the contributions of gloRE, locRE, REpre, REder, and PR in the proposed method 
and the best results are highlighted with \textbf{bold}.}\label{t4}
\centering
\tabcolsep=15pt
{\scriptsize
\begin{tabular}{c|l|cccc}
\toprule
\toprule
{\bf Dataset} & {\bf Model} &   {\bf ACC} &  {\bf NMI} &  {\bf ARI} &  {\bf F1} \\ 
\midrule
\multirow{7}{*}{\textbf{ACM}} & R$^2$FGC w/o gloRE& 92.39±0.14& 72.39±0.44& 78.72±0.37 & 92.40±0.14 \\
		 & R$^2$FGC w/o locRE & 92.21±0.16 & 72.17±0.53 & 78.53±0.43 & 92.19±0.16 \\
		 & R$^2$FGC w/o REpre & 92.13±0.23 & 71.88±0.67& 78.09±0.56 & 92.27±0.21 \\
		 & R$^2$FGC w/o REder & 92.35±0.17 & 72.34±0.55 & 78.62±0.46 & 92.41±0.18 \\
		 & R$^2$FGC w/o REpre \& REder & 92.05±0.16 & 71.31±0.49& 77.81±0.41& 92.11±0.16 \\
		 & R$^2$FGC w/o PR  & 91.93±0.18& 71.01±0.51 & 77.44±0.47 &91.95±0.18 \\
		 & \multicolumn{1}{>{\columncolor{gray!25}}l}{R$^2$FGC (Ours)}  &  \multicolumn{1}{>{\columncolor{gray!25}}l}{{\bf 92.43±0.18}} & \multicolumn{1}{>{\columncolor{gray!25}}l}{{\bf 72.42±0.53}} & \multicolumn{1}{>{\columncolor{gray!25}}l}{{\bf 78.72±0.47}} & \multicolumn{1}{>{\columncolor{gray!25}}l}{{\bf 92.45±0.18}}  \\
\midrule
\multirow{7}{*}{\textbf{AMAP}} & R$^2$FGC w/o gloRE& 81.21±0.07& 73.79±0.18 & 66.04±0.45 & 75.14±0.45  \\
		 & R$^2$FGC w/o locRE & 81.23±0.06 & 73.81±0.22 & 66.15±0.51 & 75.03±0.48  \\
		 & R$^2$FGC w/o REpre & 81.24±0.06& 73.81±0.19 & 66.20±0.42& 75.21±0.33 \\
		 & R$^2$FGC w/o REder & 80.85±0.45 & 73.06±0.69 & 66.00±0.43 & 74.88±0.46  \\
		 & R$^2$FGC w/o REpre \& REder& 80.75±0.07 & 72.87±0.23& 65.83±0.36 & 74.51±0.40 \\
		 & R$^2$FGC w/o PR  & 80.39±0.61 & 72.50±0.82& 65.53±0.67 & 74.35±0.82 \\
		 & \multicolumn{1}{>{\columncolor{gray!25}}l}{R$^2$FGC (Ours)}  & \multicolumn{1}{>{\columncolor{gray!25}}l}{{\bf 81.28±0.05}} & \multicolumn{1}{>{\columncolor{gray!25}}l}{{\bf 73.88±0.17}} & \multicolumn{1}{>{\columncolor{gray!25}}l}{{\bf 66.25±0.36}} & \multicolumn{1}{>{\columncolor{gray!25}}l}{{\bf 75.29±0.32}} \\
\midrule
\multirow{7}{*}{\textbf{CITE}} & R$^2$FGC w/o gloRE& 69.84±0.53& 44.47±0.32& 45.61±0.41 & 64.77±0.32 \\
		 & R$^2$FGC w/o locRE & 70.25±0.51 & 44.99±0.40 & 46.65±0.37 & 64.93±0.43 \\
		 & R$^2$FGC w/o REpre & 70.03±0.58 & 44.37±0.64 & 46.03±0.57 & 64.73±0.48 \\
		 & R$^2$FGC w/o REder & 68.84±0.45 & 43.06±0.47 & 44.59±0.51 & 64.37±0.38 \\
		 & R$^2$FGC w/o REpre \& REder & 68.20±0.41 & 42.97±0.32 & 44.29±0.41 & 64.30±0.29 \\
		 & R$^2$FGC w/o PR  & 69.70±0.57 & 44.36±0.43& 45.89±0.39& 64.97±0.29 \\
		 & \multicolumn{1}{>{\columncolor{gray!25}}l}{R$^2$FGC (Ours)}  & \multicolumn{1}{>{\columncolor{gray!25}}l}{{\bf 70.60±0.45}} & \multicolumn{1}{>{\columncolor{gray!25}}l}{{\bf 45.39±0.37}} & \multicolumn{1}{>{\columncolor{gray!25}}l}{{\bf 47.07±0.30}} & \multicolumn{1}{>{\columncolor{gray!25}}l}{{\bf 65.28±0.12}} \\
\midrule
\multirow{7}{*}{\textbf{DBLP}} & R$^2$FGC w/o gloRE& 80.15±0.40& 49.74±0.52 & 55.34±0.26 & 79.73±0.37 \\
		 & R$^2$FGC w/o locRE & 80.72±0.27 & 50.73±0.41& 56.23±0.52 & 80.21±0.25 \\
		 & R$^2$FGC w/o REpre & 80.82±0.25 & 50.56±0.40 & 56.04±0.52 & 80.31±0.24 \\
		 & R$^2$FGC w/o REder & 79.63±0.44 & 49.41±0.49 & 55.65±0.68 & 79.69±0.30 \\
		 & R$^2$FGC w/o REpre \& REder & 78.95±0.23 & 48.77±0.40 & 55.27±0.32 & 79.65±0.22 \\
		 & R$^2$FGC w/o PR  & 80.73±0.14& 49.59±0.26& 55.88±0.25& 80.23±0.16 \\
		 & \multicolumn{1}{>{\columncolor{gray!25}}l}{R$^2$FGC (Ours)}  & \multicolumn{1}{>{\columncolor{gray!25}}l}{{\bf 80.95±0.20}} & \multicolumn{1}{>{\columncolor{gray!25}}l}{{\bf 50.82±0.32}} & \multicolumn{1}{>{\columncolor{gray!25}}l}{{\bf 56.34±0.42}} & \multicolumn{1}{>{\columncolor{gray!25}}l}{{\bf 80.54±0.19}} \\
\midrule
\multirow{7}{*}{\textbf{HHAR}} & R$^2$FGC w/o gloRE& 88.82±0.05& 83.38±0.04 & 78.43±0.07& 89.13±0.05 \\
		 & R$^2$FGC w/o locRE & 88.87±0.05 & 83.32±0.06 & 78.45±0.07& 89.18±0.05 \\
		 & R$^2$FGC w/o REpre & 88.87±0.05 & 83.32±0.07 & 78.45±0.08 & 89.18±0.05 \\
		 & R$^2$FGC w/o REder & 88.83±0.05 & 83.37±0.04 & 78.44±0.06 & 89.15±0.05 \\
		 & R$^2$FGC w/o REpre \& REder & 88.79±0.04& 83.01±0.04& 78.05±0.05 & 88.85±0.04 \\
		 & R$^2$FGC w/o PR & 87.98±0.03 & 82.84±0.08 & 77.45±0.05 & 88.30±0.03 \\ 
		 & \multicolumn{1}{>{\columncolor{gray!25}}l}{R$^2$FGC (Ours)}  & \multicolumn{1}{>{\columncolor{gray!25}}l}{{\bf 88.91±0.05}} & \multicolumn{1}{>{\columncolor{gray!25}}l}{{\bf 83.39±0.07}} & \multicolumn{1}{>{\columncolor{gray!25}}l}{{\bf 78.52±0.08}} & \multicolumn{1}{>{\columncolor{gray!25}}l}{{\bf 89.23±0.06}} \\
\bottomrule
\bottomrule
\end{tabular}
}
\end{table*}

\begin{itemize}

\item Compared with shallow clustering method $k$-means, these deep graph clustering methods clearly show preferable performance. It indicates that the strong capability for learning representation of deep neural network methods enables exploit more meaningful information from graph-structured data for clustering. 

\item The purely AE-based methods (AE, DEC, and IDEC) perform worse than the methods combining AE and GCN (SDCN, DFCN, and AGCC) in most cases. The reason may be that the AE-based methods only leverage the attribute information to learn the latent representation, which overlooks the structure-level semantic information. Similarly, the purely GCN-based methods (GAE, VGAE, DAEGC, and ARGA) also show inferior performance than SDCN, DFCN, and AGCC in most circumstances. It indicates that integrating AE into GCN can capture the attribute and structure information more effectively from complementary views. 

\item Our method R$^2$FGC achieves the best clustering performance compared with all the baselines in terms of the four considered metrics over all the datasets. For both graph and non-graph data, our approach represents a significant improvement over the baselines. For example, compared with the best results among all the baselines, for the ACM dataset, our method relatively improves 1.68\%, 4.35\%, 5.10\%, 1.82\% on ACC, NMI, ARI, F1; for the AMAP dataset, our method improves 5.72\%, 6.75\%, 11.55\%, 5.18\% 
on ACC, NMI, ARI, F1; for the DBLP dataset, our method improves 6.51\%, 16.29\%, 19.87\%, 6.39\% 
on ACC, NMI, ARI, F1, respectively. 

\item The reasons for the superiority of our method R$^2$FGC are that 
a) R$^2$FGC extracts the inherent relational information based on AE and GAE from both local and global views under augmentation, which allows for better exploration of both attribute and structure information; 
b) Under augmentation, R$^2$FGC preserves the consistent relationship among the nodes but not the latent representations, which expects to learn more essential representations of the semantic information; 
c) R$^2$FGC decreases the redundant relation among the nodes for learning discriminative and meaningful representations, which can better serve the graph clustering; 
d) R$^2$FGC couples AE and GAE together in the representation fusion mechanism to fully integrate and refine the attribute and structure information; 
e) R$^2$FGC also brings the propagation regularization to mitigate the possible over-smoothing problem caused by GAE to promote the clustering performance. 
With the addition of relation extraction, relation preservation and de-redundancy strategies, R$^2$FGC outperforms all the baselines upon the fusion mechanism of AE and GAE and the regularization method of alleviating over-smoothing. 

\end{itemize}


\subsection{Ablation Study (RQ2)}

In this section, to further investigate the validity of our proposed method, we conduct some ablation experiments to study the contribution of each component of R$^2$FGC. We mainly focus on the influence of global-view relation extraction (gloRE), local-view relation extraction (locRE), relation preservation (REpre), relation de-redundancy(REder), and propagation regularization (PR). In addition, we make some discussion on the proposed global sampling strategy. 

{\bf Effects of gloRE and locRE.} In the relation extraction module, we explore the inherent relation from both global and local views. The former view learns the global relation of the nodes and the latter concerns the neighbor relation. We perform some ablation experiments to verify the respective effectiveness of the global- and local-view strategies. Specifically, we consider the following two cases: 
\begin{itemize}
\item R$^2$FGC w/o gloRE: R$^2$FGC without considering the global-view relation extraction;
\item R$^2$FGC w/o locRE: R$^2$FGC without considering the local-view relation extraction;
\end{itemize}

The corresponding results are displayed in Table \ref{t4}. From the comparison of R$^2$FGC w/o gloRE and R$^2$FGC w/o locRE, for the ACM dataset, local-view relation extraction has a greater effect than the global-view one on clustering in terms of ACC, NMI, ARI, and F1, while for the CITE and DBLP datasets, the global-view relation extraction may have a more prominent contribution. As for the AMAP and HHAR datasets, R$^2$FGC w/o gloRE and R$^2$FGC w/o locRE show close metric values, which indicates that global- and local-view extractions almost play equal roles. Moreover, R$^2$FGC consistently shows better performance than R$^2$FGC w/o gloRE and R$^2$FGC w/o locRE over the five considered datasets. Hence, these results illustrate that both views are necessary and important for achieving good clustering performance.

{\bf Effects of gloRE, locRE, and PR.} Additionally, relation preservation is used to learn the effective representations by preserving the consistent relation information, whereas the relation de-redundancy conduces to reduce the confusing information, which benefits obtaining discriminative embeddings. Moreover, we adopt propagation regularization to relieve the over-smoothing issue. Hence, we also explore their respective efficiencies in the ablation experiments, i.e., four cases are considered as follows: 
\begin{itemize}
\item R$^2$FGC w/o REpre: R$^2$FGC without considering the relation preservation;
\item R$^2$FGC w/o REder: R$^2$FGC without considering the relation de-redundancy; 
\item R$^2$FGC w/o REpre \& REder: R$^2$FGC without both relation preservation and de-redundancy; 
\item R$^2$FGC w/o PR: R$^2$FGC without adopting the propagation-regularization trick. 
\end{itemize}

The corresponding results are also shown in Table \ref{t4}. Comparing R$^2$FGC w/o REpre with R$^2$FGC w/o REder, it is observed that relation preservation outperforms relation de-redundancy for the ACM dataset; relation de-redundancy shows more significant power to improve the clustering performance for the AMAP, CITE, and DBLP datasets; these two strategies have almost equal impact on the HHAR dataset. Moreover, by contrasting with R$^2$FGC w/o REpre \& REder, both R$^2$FGC w/o REpre and R$^2$FGC w/o REder give better clustering results with higher metric values, which implies that both relation preservation and relation de-redundancy possess the capability to promote the effect of graph clustering. In addition, by comparing R$^2$FGC w/o PR and R$^2$FGC w/o REpre \& REder, for the ACM, AMAP, and HHAR datasets, the over-smoothing issue has a more significant impact on clustering performance, whereas, for the CITE and DBLP datasets, the relation extraction is more important for good performance. For example, R$^2$FGC on the ACM dataset has 1.99\% relative improvement over R$^2$FGC w/o PR in terms of NMI; R$^2$FGC on the CITE dataset obtains 6.28\% improvement over R$^2$FGC w/o REpre \& REder in terms of ARI. Hence, these results demonstrate that all of the proposed components in R$^2$FGC are efficient for reaching informative representation and good performance for graph clustering.

\begin{figure}[!t]
\centering
\subfigure[ACC]{
\includegraphics[width=0.22\textwidth]{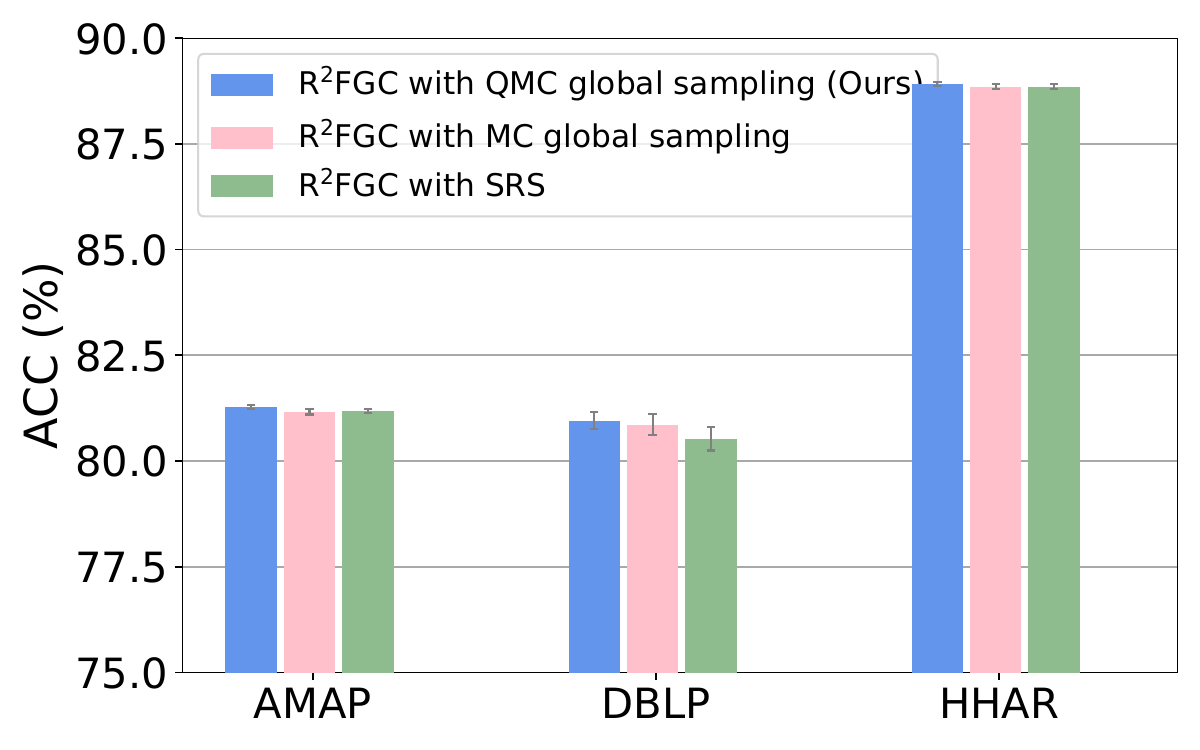}
}
\subfigure[NMI]{
\includegraphics[width=0.22\textwidth]{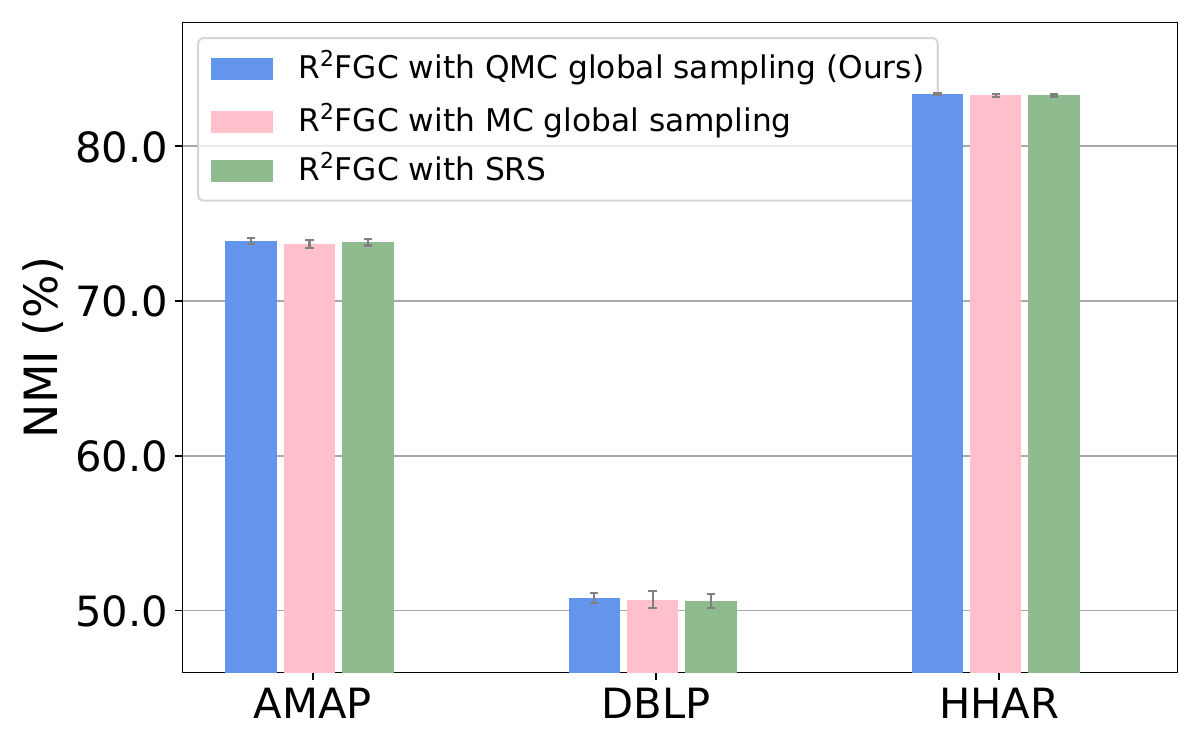}
}
\subfigure[ARI]{
\includegraphics[width=0.22\textwidth]{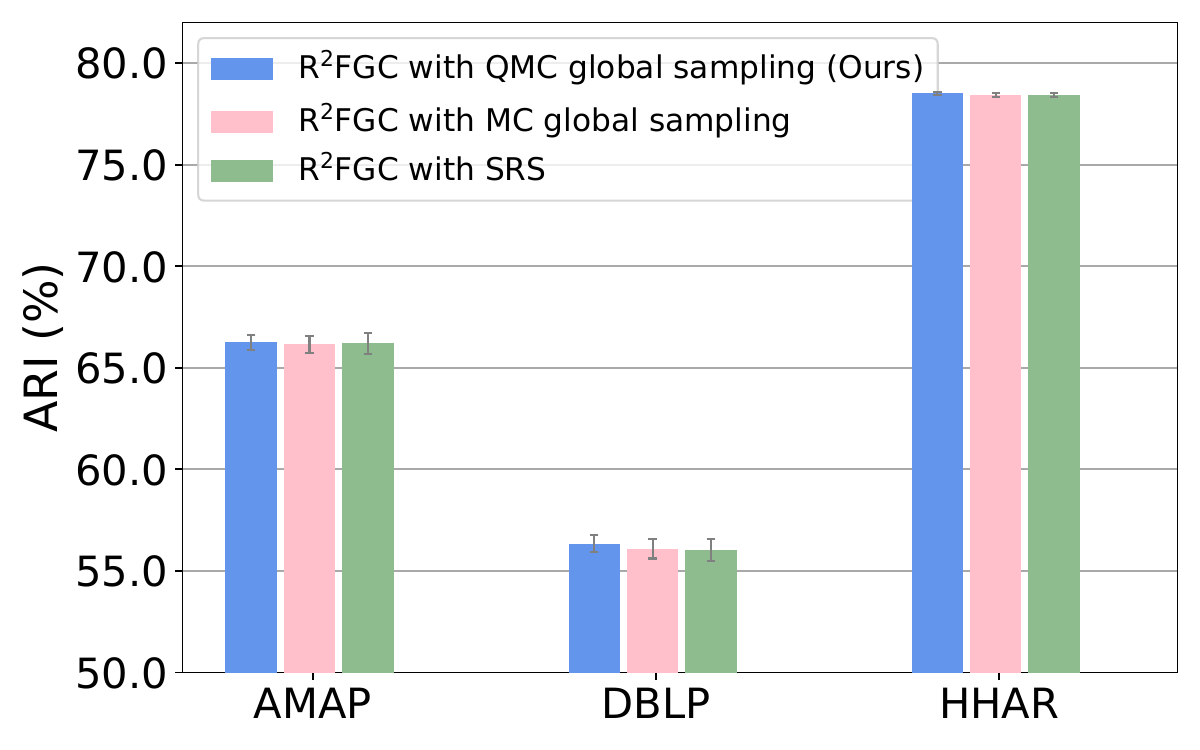}
}
\subfigure[F1]{
\includegraphics[width=0.22\textwidth]{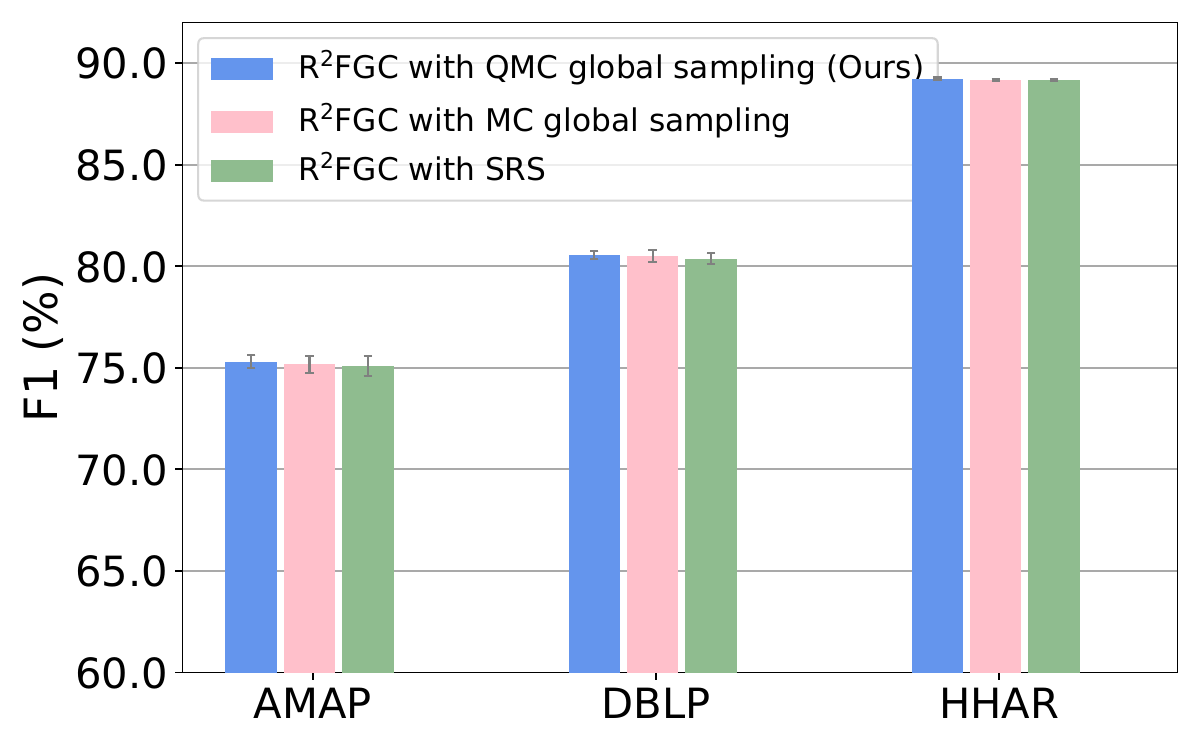}
}
\caption{The performance comparisons $w.r.t.$ different global sampling strategies on the AMAP, DBLP, and HHAR datasets.}
\label{dissam}  
\end{figure}

{\bf Discussion on Global Sampling Strategy.} To further illustrate the effectiveness of the QMC inverse degree-weighted distribution sampling for extracting global anchors, we perform experiments to compare it with two MC cases on the AMAP, DBLP, and HHAR datasets, i.e., we consider the following three cases:
\begin{itemize}
\item R$^2$FGC with QMC global sampling (Ours): R$^2$FGC with considering QMC inverse degree-weighted distribution sampling, i.e., the low-discrepancy point set is used in the multinomial sampling; 
\item R$^2$FGC with MC global sampling: R$^2$FGC with considering MC inverse degree-weighted distribution sampling, i.e., the uniform random numbers are used;  
\item R$^2$FGC with SRS: R$^2$FGC with considering simple random sampling for drawing global anchors. 
\end{itemize} 
The results are depicted in Figure \ref{dissam}. Comparing the three strategies, R$^2$FGC with QMC global sampling shows better performance over the three considered datasets in terms of the average ACC, NMI, ARI, and F1 scores. Moreover, R$^2$FGC with MC global sampling outperforms R$^2$FGC with SRS, which implies that inverse degree-weighted distribution sampling is indeed effective to avoid poor representations. In addition, from the error bars in Figure \ref{dissam}, we can also find that R$^2$FGC with QMC global sampling leads to smaller variances for the metric values, which benefits from the high convergence rate of the QMC sampling strategy. The sampled global anchor set is a better representation of the target distribution, which motivates the subsequent representation learning to have a better and more stable performance. In this way, our proposed sampling method guarantees good robustness to relation extraction and thus to clustering performance.

\begin{figure*}[!t]
\centering
\subfigure[ACM]{
\includegraphics[width=0.23\textwidth]{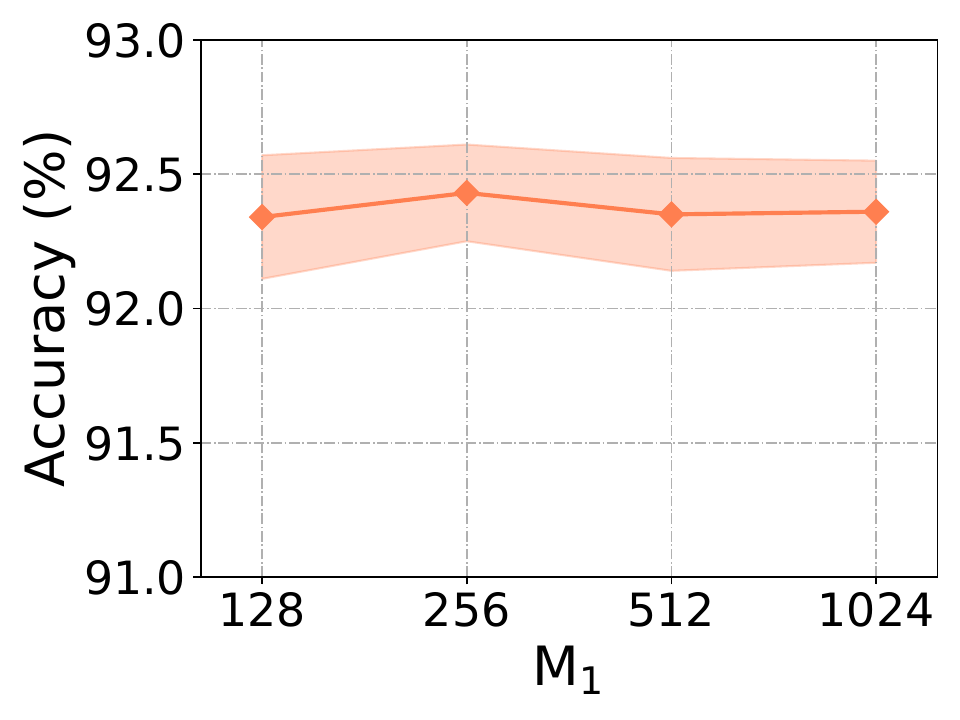}
\includegraphics[width=0.23\textwidth]{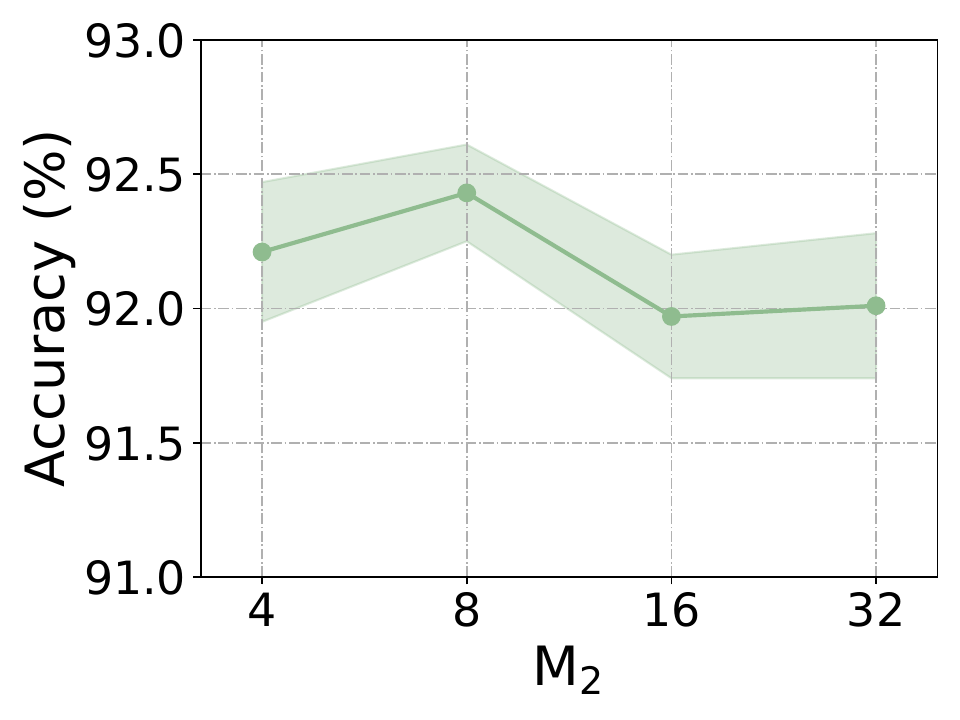}
}
\subfigure[AMAP]{
\includegraphics[width=0.23\textwidth]{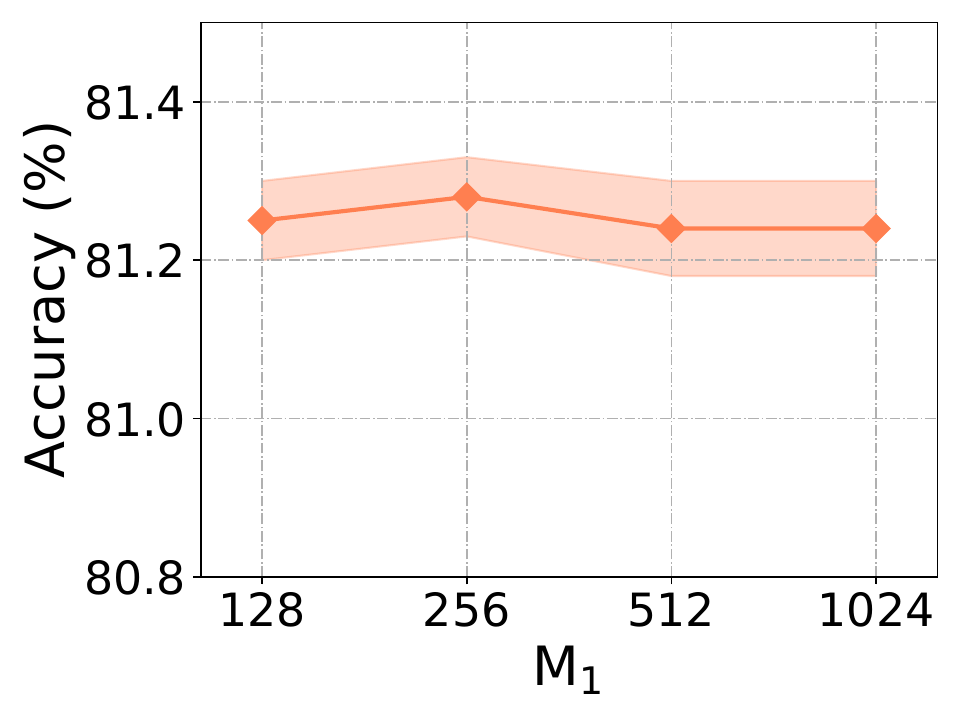}
\includegraphics[width=0.23\textwidth]{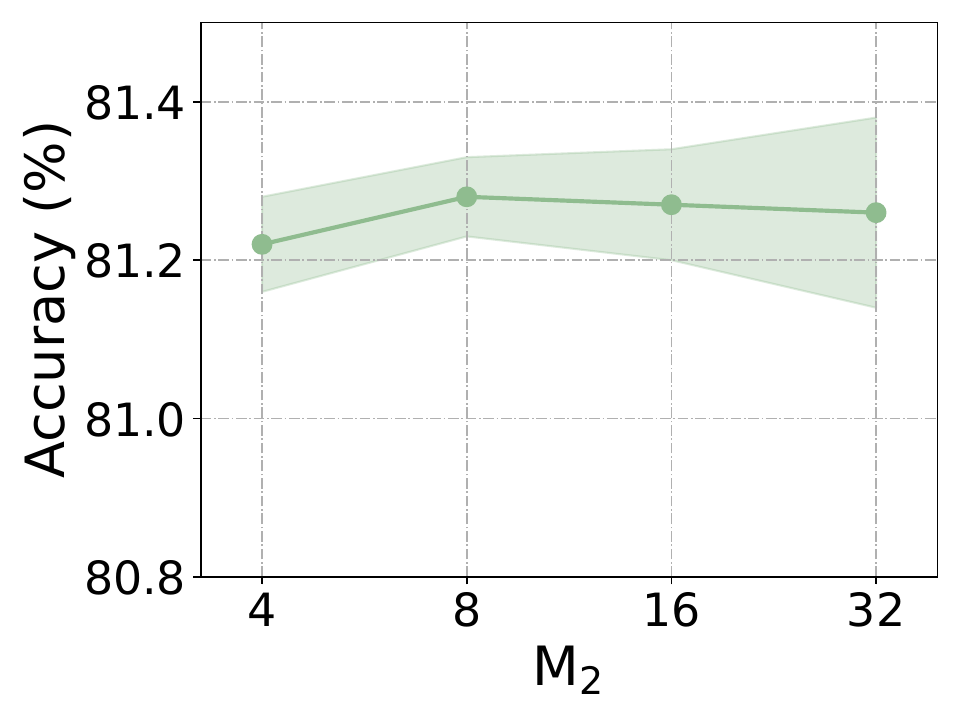}
}
\subfigure[CITE]{
\includegraphics[width=0.23\textwidth]{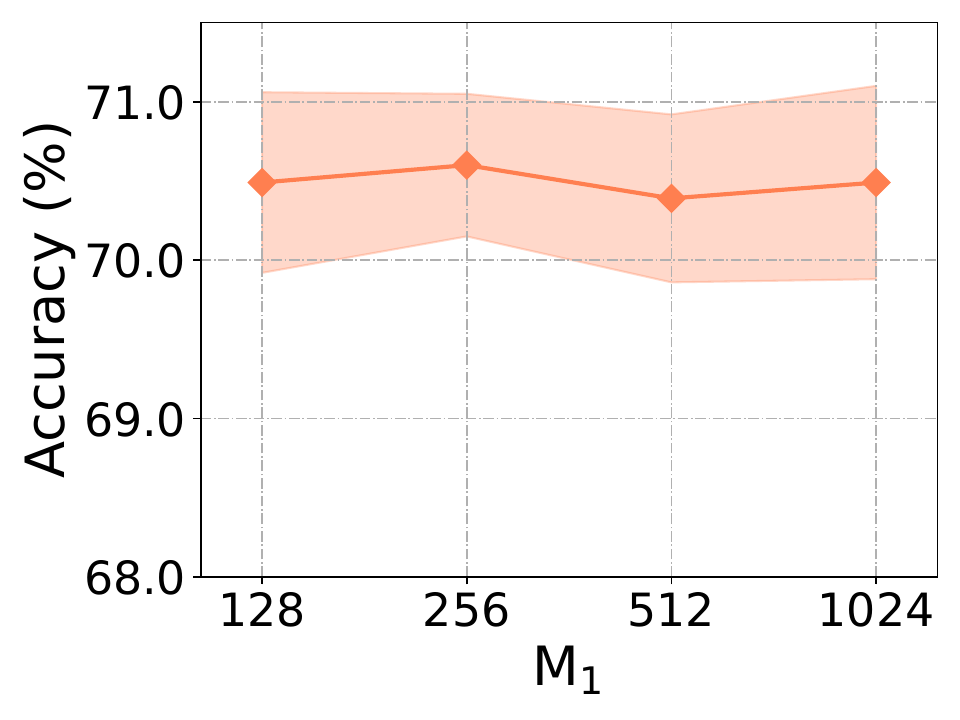}
\includegraphics[width=0.23\textwidth]{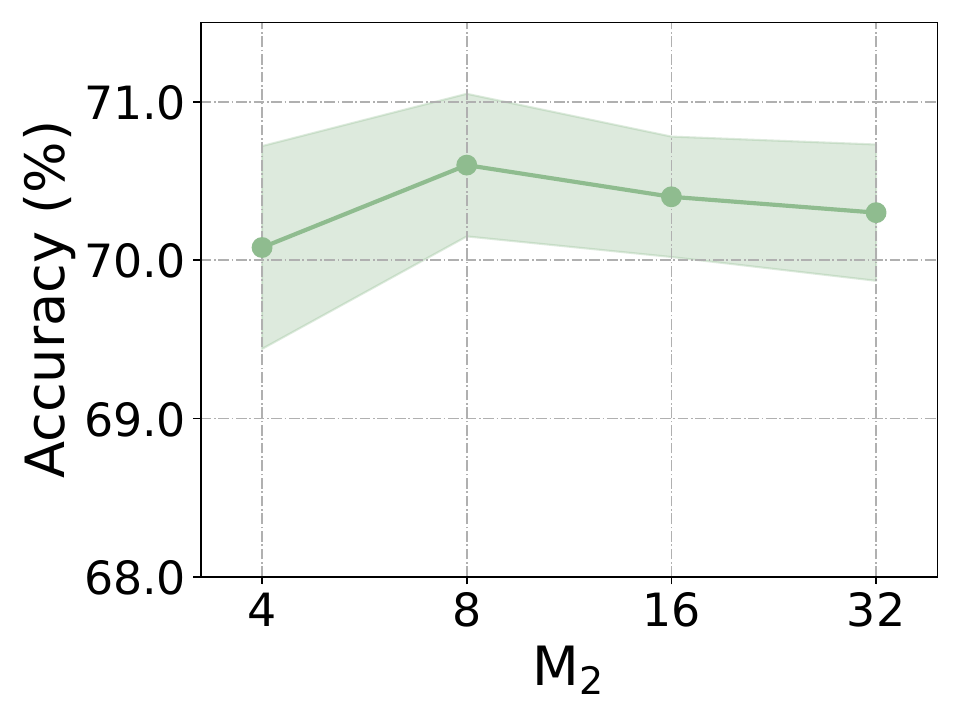}
}
\subfigure[HHAR]{
\includegraphics[width=0.23\textwidth]{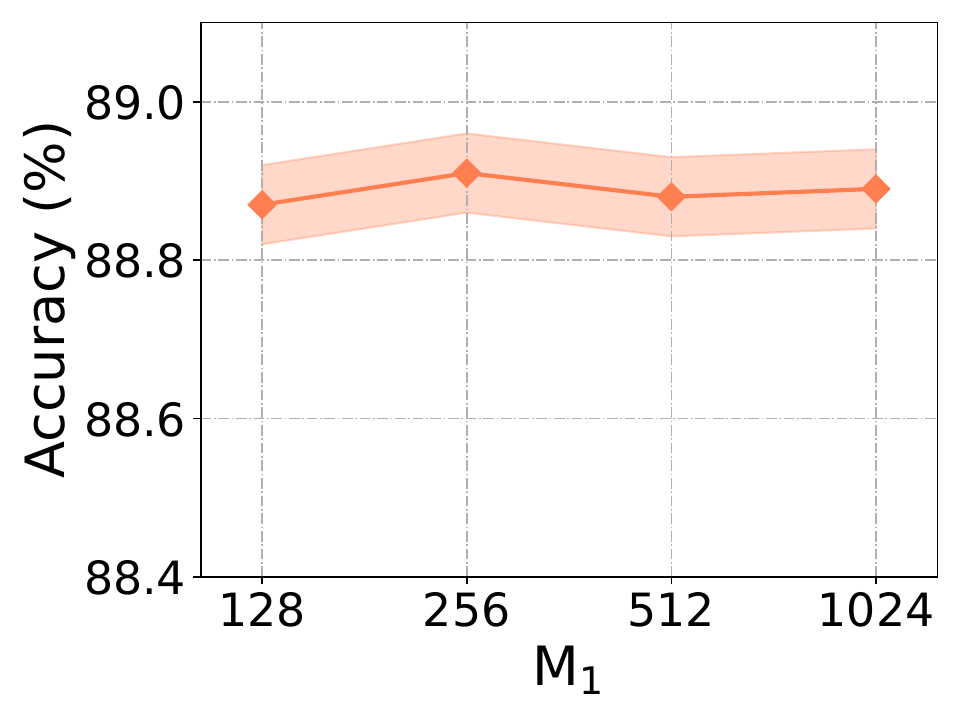}
\includegraphics[width=0.23\textwidth]{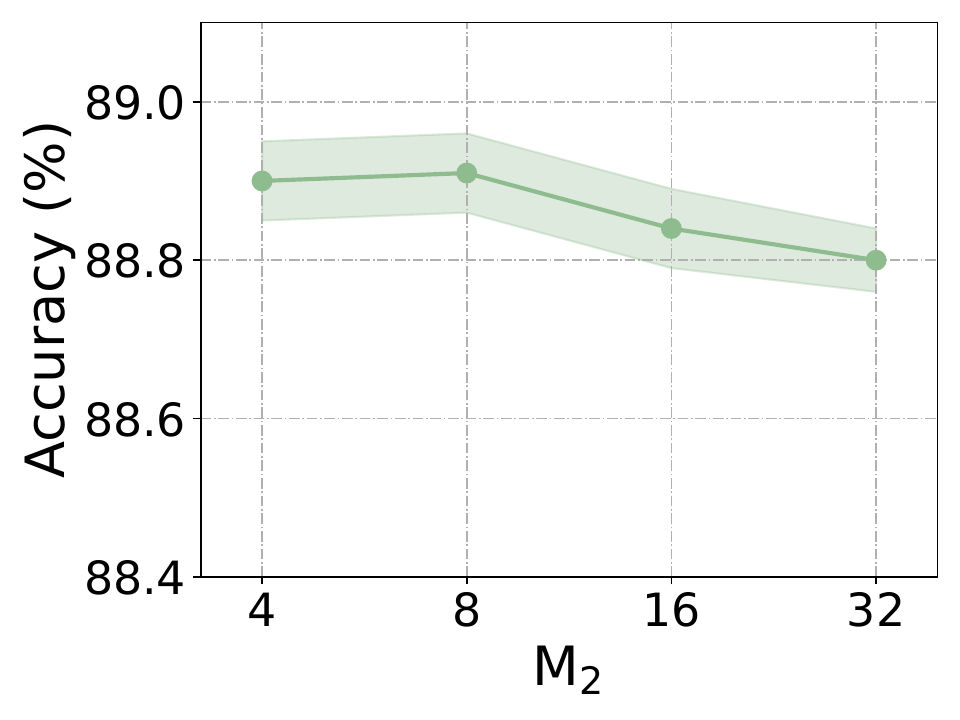}
}
\caption{The performance comparisons $w.r.t.$ different amounts of global anchors $M_1$ and local anchors $M_2$ on the ACM, AMAP, CITE, and HHAR datasets.}
\label{ps}  
\end{figure*}

\begin{figure*}[!t]
\centering
\subfigure[ACM]{
\includegraphics[width=0.23\textwidth]{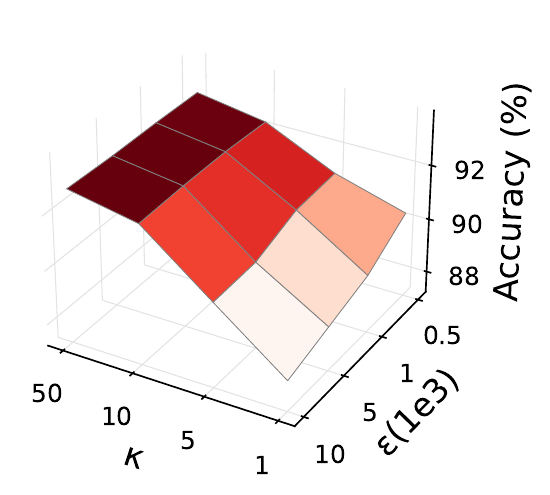}
}
\subfigure[AMAP]{
\includegraphics[width=0.23\textwidth]{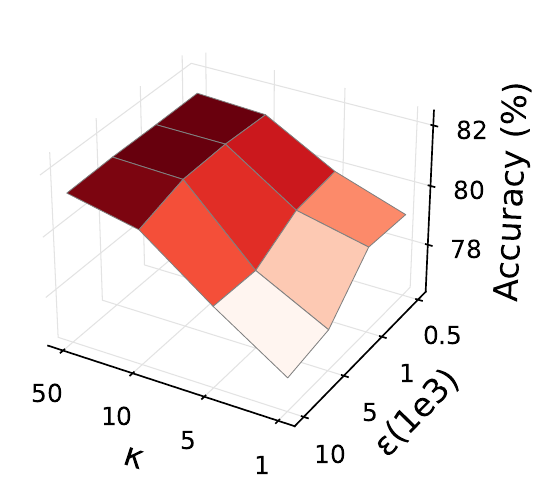}
}
\subfigure[CITE]{
\includegraphics[width=0.23\textwidth]{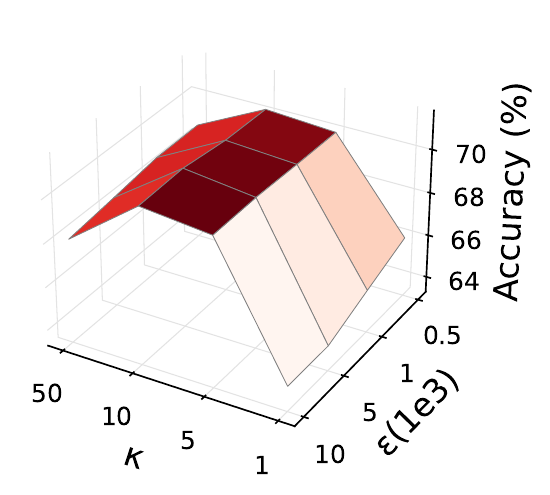}
}
\subfigure[DBLP]{
\includegraphics[width=0.23\textwidth]{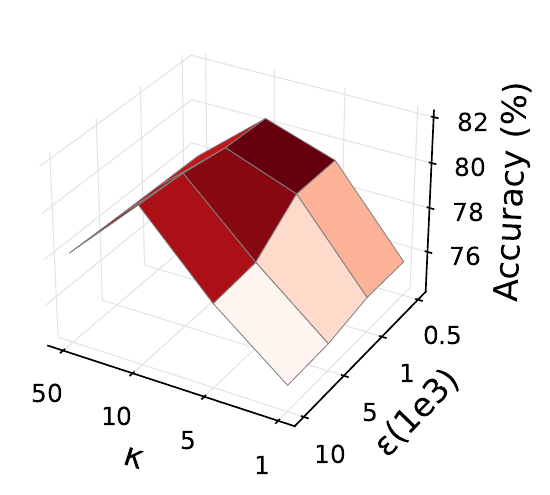}
}
\caption{The performance comparisons $w.r.t.$ different loss weight parameters $\kappa$ and $\epsilon$ on the ACM, AMAP, CITE, and DBLP datasets.}
\label{weightsensi}  
\end{figure*}

\subsection{Parameter Sensitivity Analysis (RQ3)}

In this section, we examine the sensitivity of the proposed R$^2$FGC to the hyper-parameters. For the global- and local-view relation extraction in Section \ref{relation}, we need to pre-define the numbers of the global and local anchors $M_1$ and $M_2$ for sampling. Hence, we investigate the effect of varying $M_1$ and $M_2$ on ACM, AMAP, CITE, and HHAR datasets. For each dataset, we consider $M_1=\{128,256,512,1024\}$ and $M_2=\{4,8,16,32\}$. When $M_1$ is varied, we fix $M_2$ to its optimal setting as in Section \ref{es}, and vice versa. Additionally, we explore the impact of two loss weight parameters $\epsilon$ and $\kappa$ on ACM, AMAP, CITE, and DBLP datasets. We vary $\epsilon$ across $\{5{\rm e}2,1{\rm e}3,5{\rm e}3,1{\rm e}4\}$ and $\kappa$ across $\{1,5,10,50\}$. The results are depicted in Figures \ref{ps} and \ref{weightsensi}, respectively.

{\bf Performance of Different Amounts of Global Anchors.} From Figure \ref{ps}, it can be seen that the average accuracies for the considered datasets are relatively stable as $M_1$ changes. It may be due to that with the QMC multinomial sampling, the drawn anchors can well mimic the defined inverse degree-weighted distribution even if $M_1$ is small. It helps to solve the problem caused by varying qualities of the learned representations for the nodes with different degrees. On the other hand, we draw different samples in different training epochs based on the randomization strategy, which increases the diversity of the samples to catch a broad relationship, even with a small number of global anchors. Therefore, the clustering performance is robust to the number of global anchors based on the proposed sampling strategy.

{\bf Performance of Different Amounts of Local Anchors.} As for $M_2$, it can be found that on the four datasets, as $M_2$ increases, it promotes the clustering performance first and then shows a weakening tendency. The possible reason may be that small $M_2$ cannot well collect the neighboring information, whereas large $M_2$ may absorb nodes involved in other clusters, which can disturb the extraction of local relation. Hence, a moderate number of local anchors is preferable and a well-designed deterministic sampling is desirable to avoid the intake of inconsistent information from other nodes. 

{\bf Performance of Different Amounts of Loss Weights.} 
As shown in Figure~\ref{weightsensi}, when $\kappa$ is small, increasing $\epsilon$ leads to a decrease in model performance. This is because large $\epsilon$ enhances the information aggregation ability of the nodes, which is equivalent to a deep GCN and thus increases the risk of over-smoothing, while small $\kappa$ means a low self-supervision ability, which results in poor cohesion and insufficient discrimination in node representations. 
As the increase of $\kappa$, better representation cohesion achieves, and increasing $\epsilon$ appropriately is promising to improve the performance by balancing the strength of neighbor aggregation in GCN and the weakness of over-smoothness. 
However, when $\epsilon$ becomes excessively large, there may be a slight decline in performance due to the over-smoothing issue on some datasets. In addition, when $\epsilon$ is fixed, increasing $\kappa$ results in an increasing trend in model performance on ACM and AMAP datasets. However, on CITE and DBLP datasets, excessively large $\kappa$ leads to a decreasing trend. One possible reason is that CITE and DBLP represent citation networks and author networks, respectively, where different articles and individuals may belong to distinct disciplines or communities. Forcing strong cohesion in these cases may lead to suboptimal results. 
Overall, we recommend to set $\kappa$ around 10 and $\epsilon$ around 5e3 for satisfying performance. When dealing with a new dataset, a small-scale hyper-parameter tuning around the recommended values is needed due to the dataset's specific characteristics.

\subsection{Empirical Convergence Analysis (RQ4)}

\begin{figure}[!t]
\centering
\subfigure[AMAP]{
\includegraphics[width=0.22\textwidth]{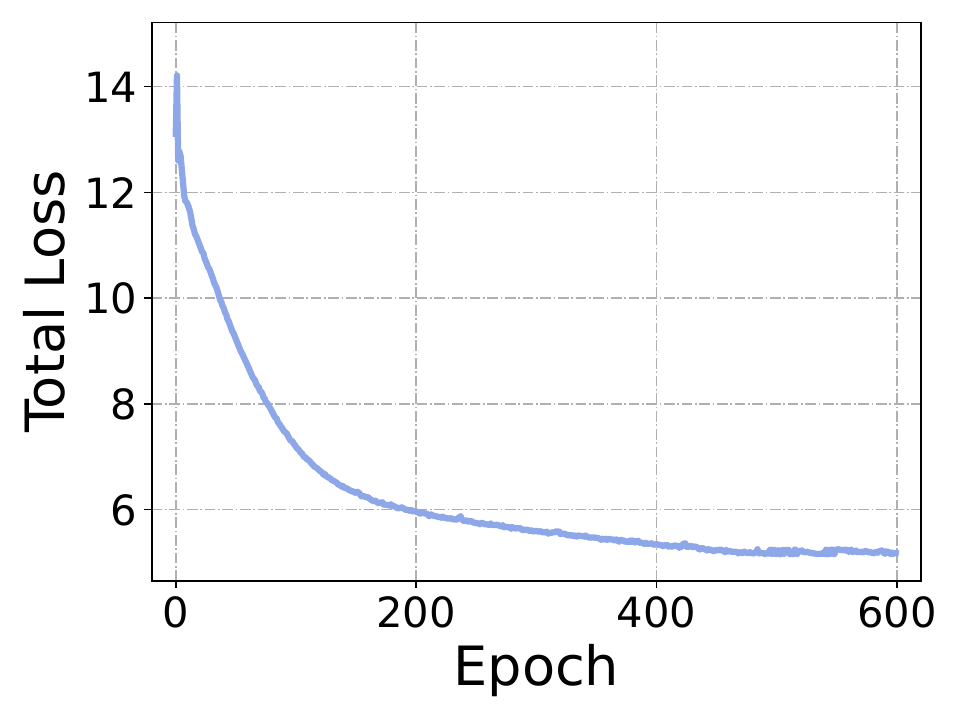}
}
\subfigure[HHAR]{
\includegraphics[width=0.22\textwidth]{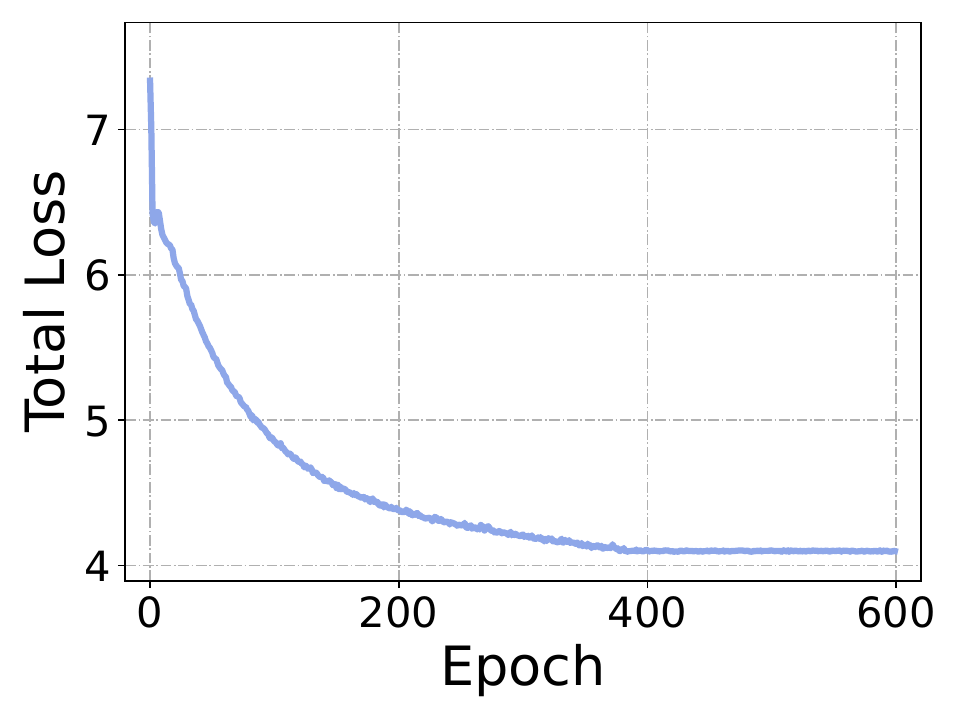}
}
\caption{The curves of the training loss against the number of epochs on the AMAP and HHAR datasets.}
\label{lossconverge}  
\end{figure}

In this section, we analyze the convergence of our proposed method R$^2$FGC, the curves of the training losses are shown in the Figure~\ref{lossconverge}. It can be observed that our method demonstrates graceful convergence across different datasets AMAP and HHAR. The reason behind this can be attributed to our pre-training learning based on AE and GAE, which provides us with a well-initialized representation. As a result, the initial loss optimization has the correct gradient direction, leading to a rapid decrease in loss. Additionally, our method effectively maintains the relational similarity between nodes in the graph while reducing the redundancy of learned representations. This allows the learned representations to possess highly rich semantic information and strong discriminative capabilities. It enables similar nodes closer to each other while better distinguishing unrelated nodes, facilitating the formation of clusters. This also motivates the training objective to converge to lower value, leading to better clustering performance. 

\subsection{Analysis of Over-smoothing Issue (RQ5)}

\begin{figure}[!t]
\centering
\subfigure[AMAP]{
\includegraphics[width=0.23\textwidth]{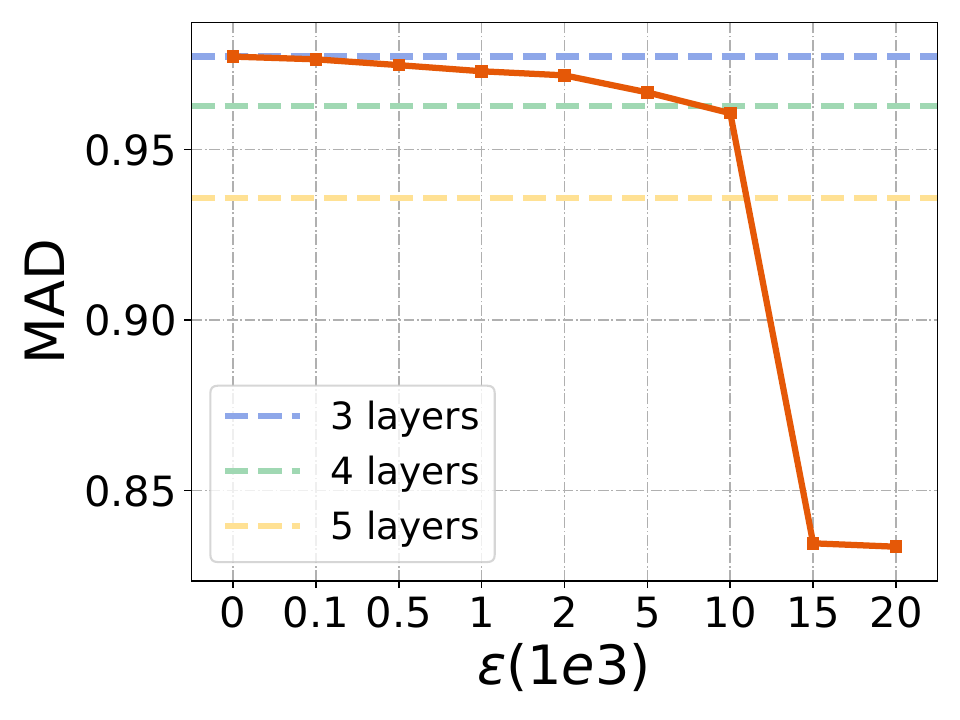}
\includegraphics[width=0.23\textwidth]{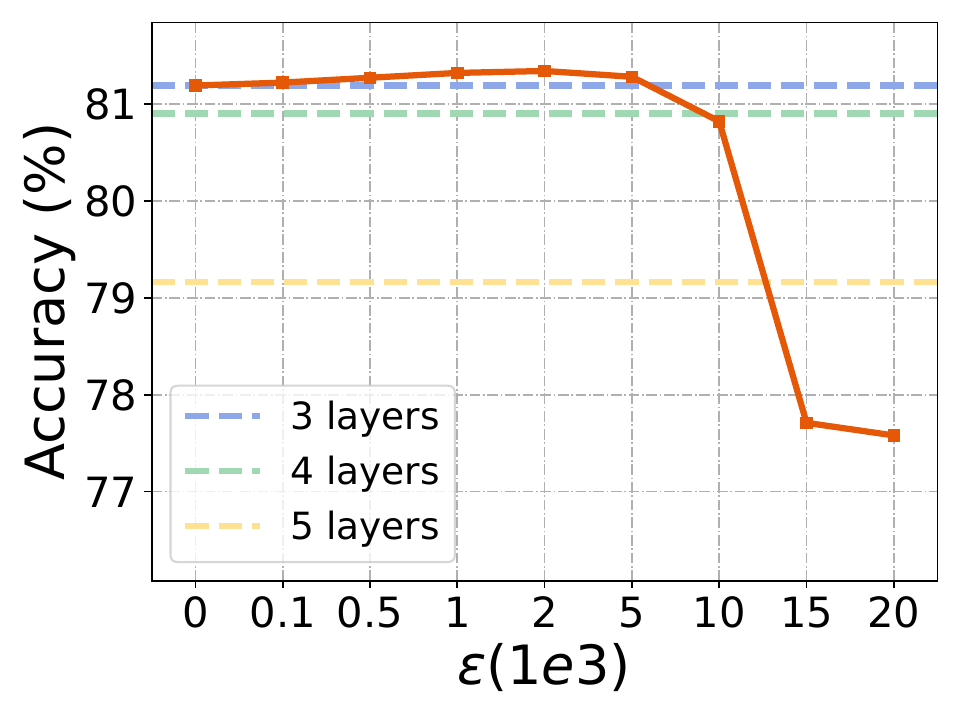}
}
\subfigure[DBLP]{
\includegraphics[width=0.23\textwidth]{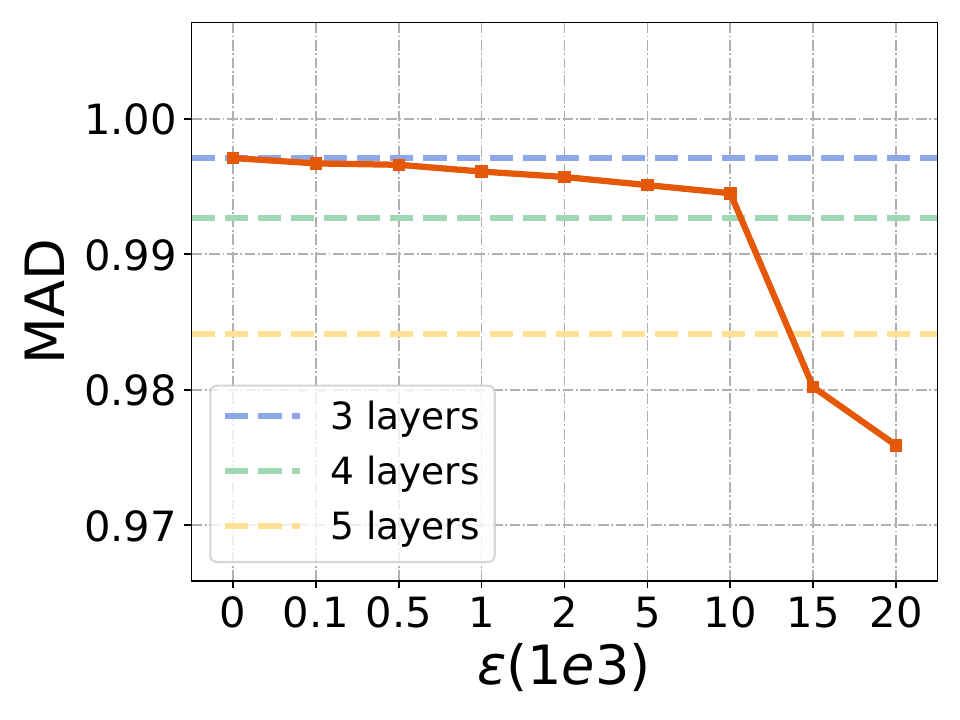}
\includegraphics[width=0.23\textwidth]{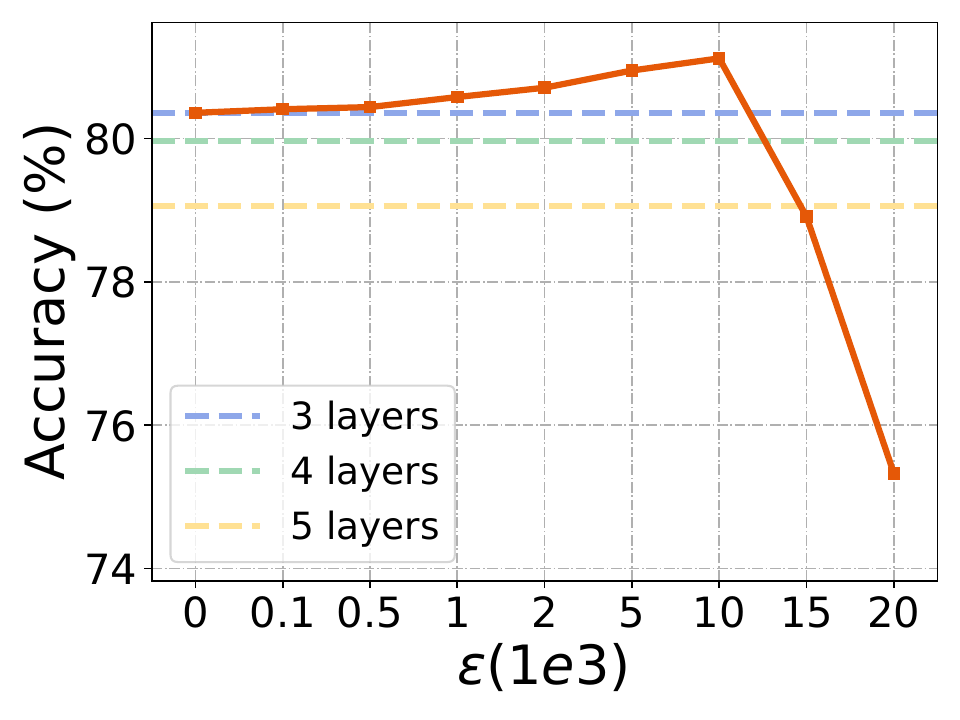}
}
\caption{The comparisons of the mean average distance (MAD) and clustering accuracy $w.r.t$ different GCN layers in GAE encoder and regularization parameter $\epsilon$ on the AMAP and DBLP datasets.}
\label{oversmooth}  
\end{figure}

To verify the superiority of the proposed propagation-regularization loss in alleviating over-smoothing issue, we compare the effects of different GCN layers in GAE encoder and different values of $\epsilon$ by mean average distance (MAD) and clustering performance (i.e., accuracy). MAD reflects the smoothness of node representations by calculating the mean of the average cosine distance between the nodes and other nodes~\cite{chen2020measuring}. A smaller MAD indicates a higher global smoothness. The analysis results are shown in Figure \ref{oversmooth}.

It can be observed that as the number of GCN layers in GAE encoder increases, indicated by the dashed line in the figure, both MAD and accuracy exhibit a decreasing trend. It implies that larger GCN layers cause nodes to immensely absorb information from farther neighbors and thus can lead to indistinguishable node representations, exacerbating the over-smoothing issue and resulting in performance degradation. 
On the other hand, our proposed propagation-regularization loss shows a slight decrease in MAD and a gradual increase in clustering performance as $\epsilon$ increases to a certain value (e.g., 2e3 for AMAP, 1e4 for DBLP). This suggests that our propagation-regularization loss is equivalent to simulating a GCN of a fractional layer, which possesses the capability to ease the increase of smoothness and meanwhile enhance the expressive power of node representations, thereby promoting the clustering performance. 
However, when $\epsilon$ is particularly large, both MAD and accuracy decrease sharply. This is because excessively large $\epsilon$ amplifies the risk of over-smoothness. Therefore, selecting an appropriate weight is crucial in balancing node expressiveness and the over-smoothing issue.

\subsection{Visualization of Clustering Results (RQ5)}

\begin{figure*}[!t]
\centering
\subfigure[Raw data]{
\includegraphics[width=0.23\textwidth]{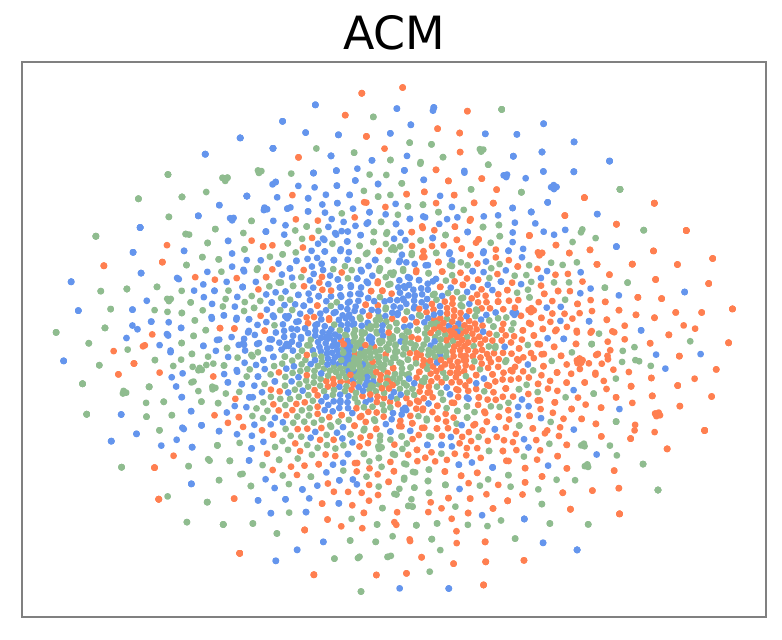}
\includegraphics[width=0.23\textwidth]{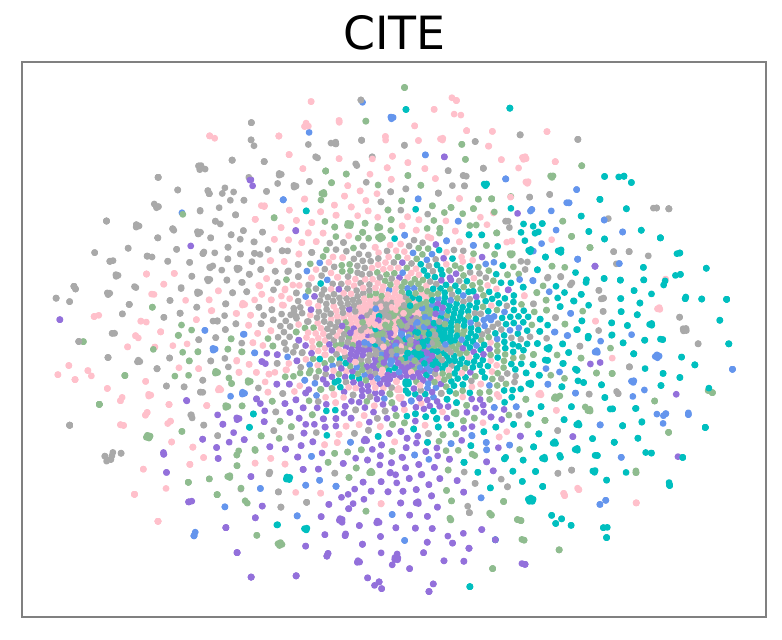}
\includegraphics[width=0.23\textwidth]{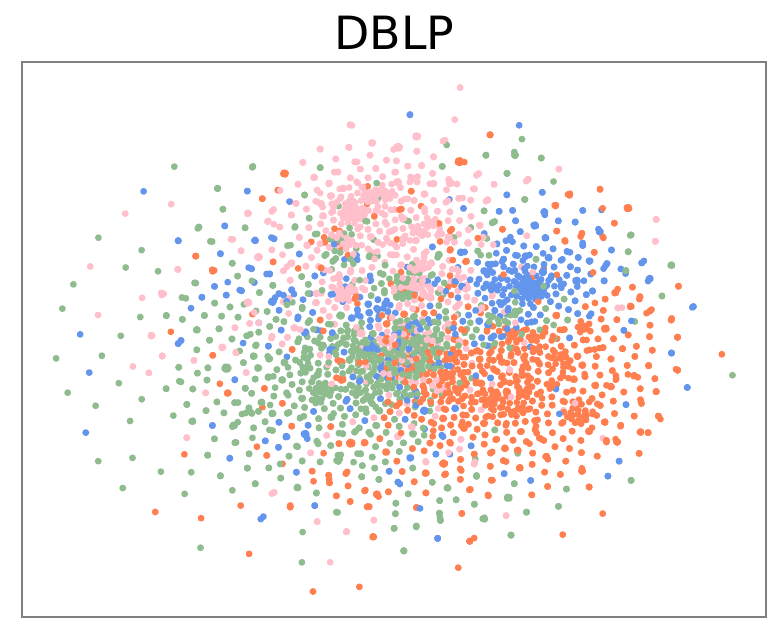}
\includegraphics[width=0.23\textwidth]{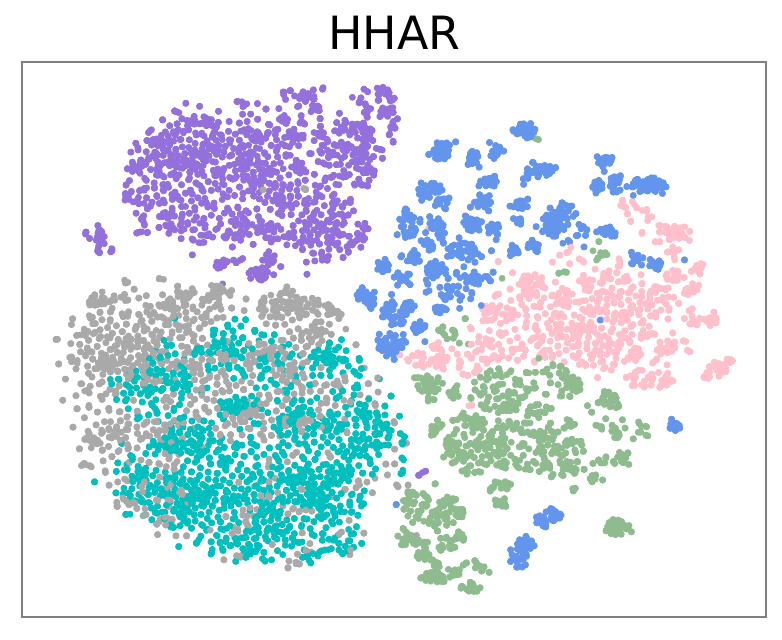}
}
\subfigure[DFCN]{
\includegraphics[width=0.23\textwidth]{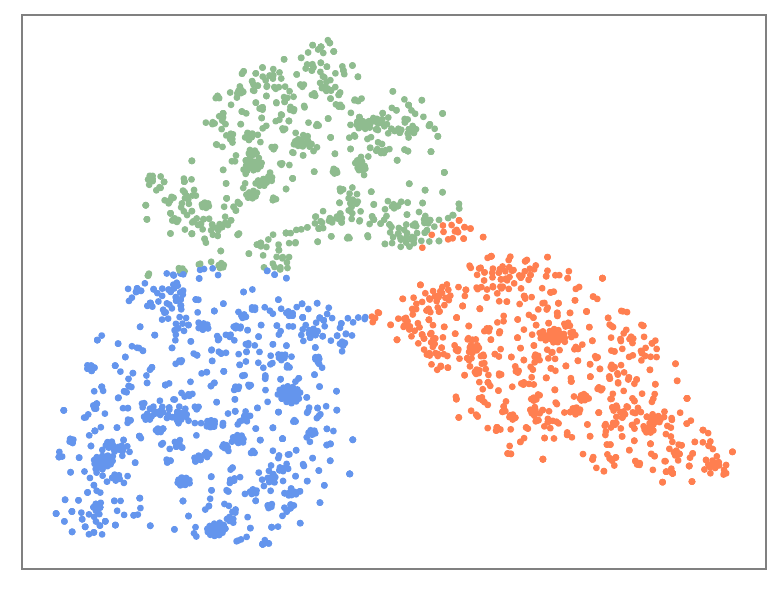}
\includegraphics[width=0.23\textwidth]{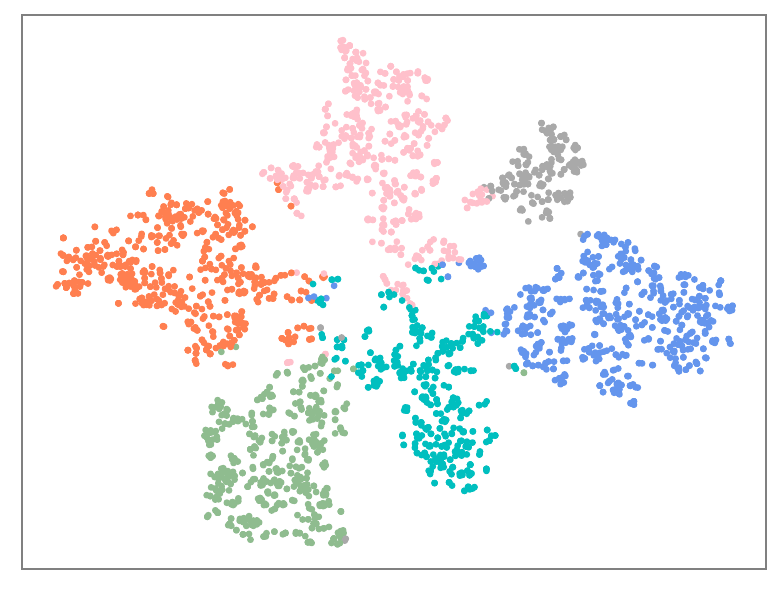}
\includegraphics[width=0.23\textwidth]{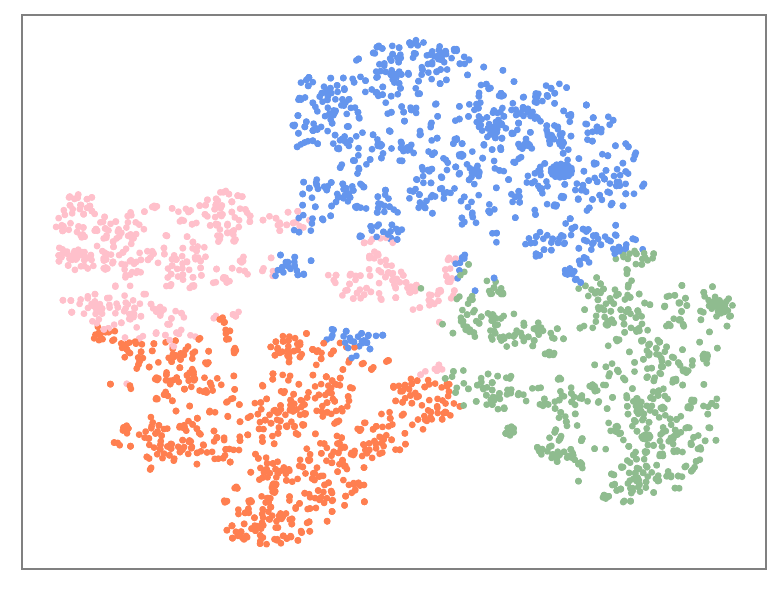}
\includegraphics[width=0.23\textwidth]{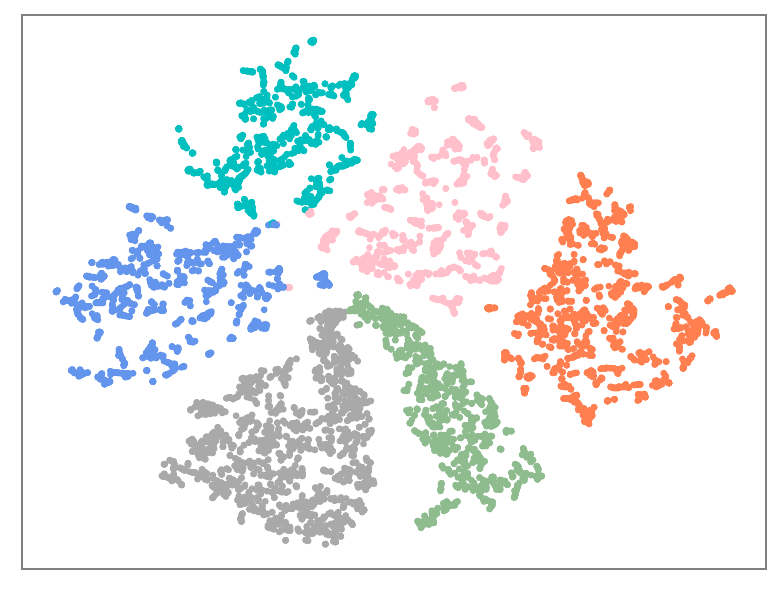}
}
\subfigure[R$^2$FGC]{
\includegraphics[width=0.23\textwidth]{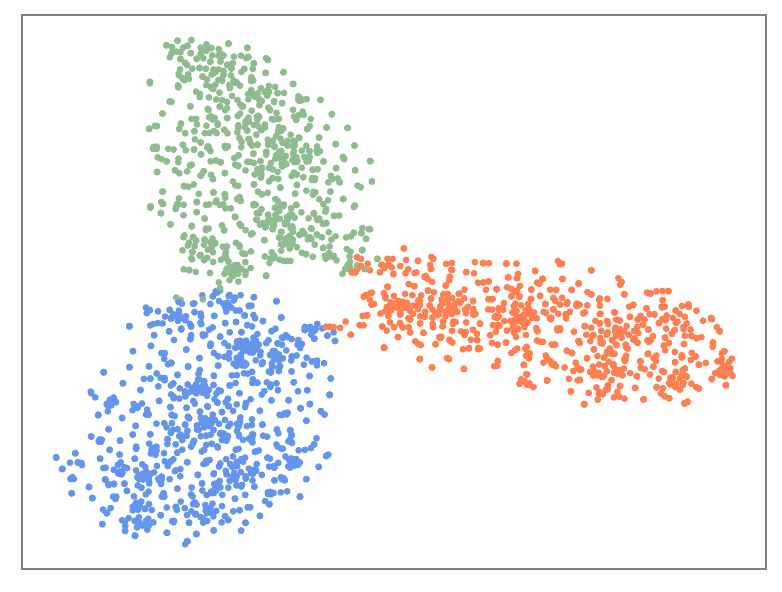}
\includegraphics[width=0.23\textwidth]{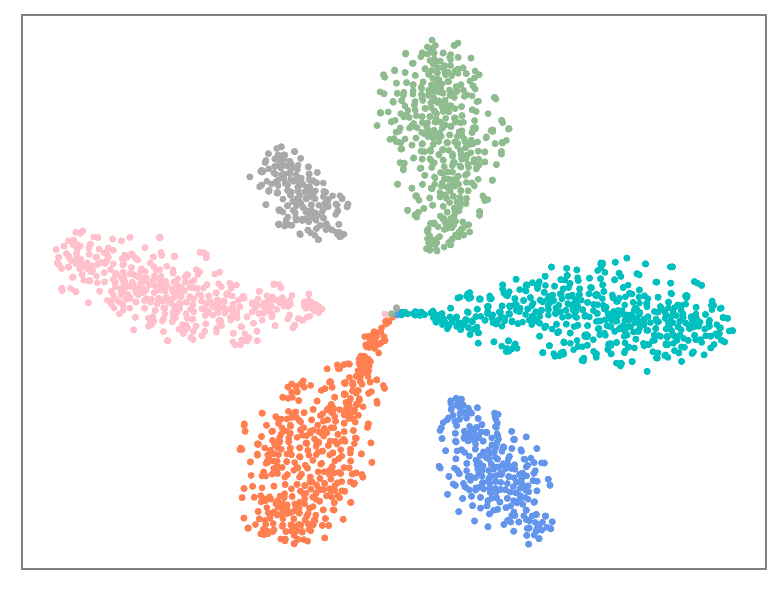}
\includegraphics[width=0.23\textwidth]{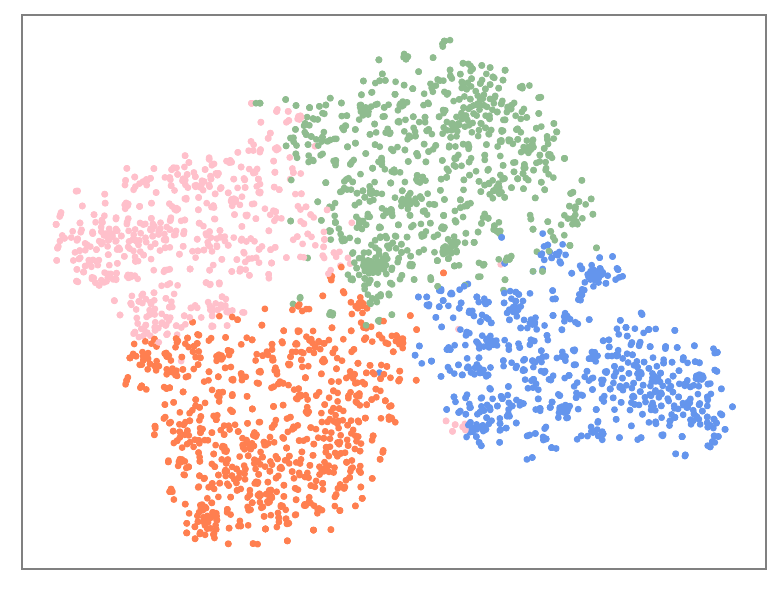}
\includegraphics[width=0.23\textwidth]{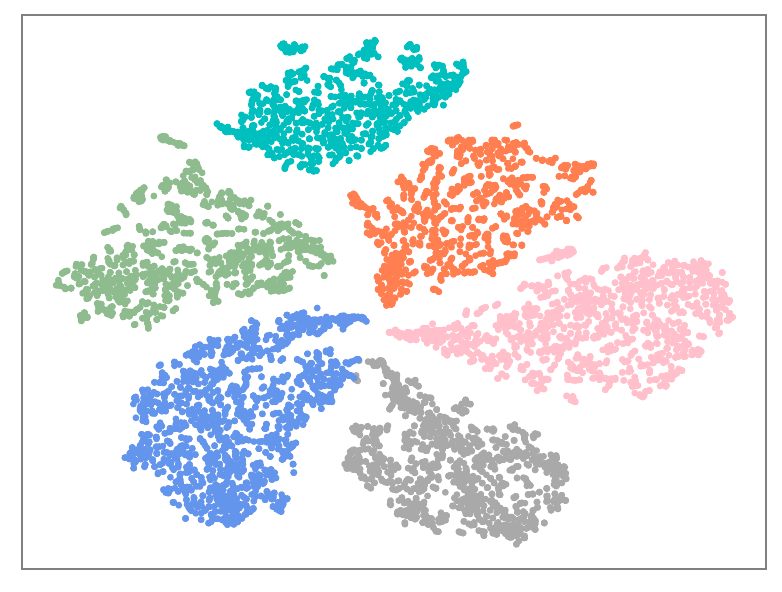}
}
\caption{The $t$-SNE visualizations on the ACM, CITE, DBLP, and HHAR datasets. The first, middle, and last rows correspond to the distributions of the embeddings from raw data, DFCN, and our proposed R$^2$FGC, respectively.}
\label{tsne}  
\end{figure*}

To visually verify the validity of our proposed R$^2$FGC, we plot 2D $t$-distributed stochastic neighbor embedding ($t$-SNE) visualizations\cite{van2008visualizing} for the learned representations on the ACM, CITE, DBLP, and HHAR datasets. We compare the $t$-SNE visualizations of the embeddings resulting from R$^2$FGC with those from the raw data and DFCN (the best method among the baselines in Section \ref{RQ1}) to enable a visual comparison. The plots are shown in Figure \ref{tsne}. The results of $t$-SNE on the four raw data clearly have poor separability for different clusters. Compared with the raw data, more distinguishing visualizations in R$^2$FGC and DFCN demonstrate that deep graph clustering methods indeed make great performance improvements. Comparing R$^2$FGC with DFCN, the latent representations obtained by our method R$^2$FGC show better separability for different clusters, where the samples from the same cluster have better aggregation and those from different clusters have a bigger gap. Such a phenomenon illustrates that our proposed method learns more discriminative representations and produces more effective cluster assignments compared with state-of-the-art methods.

\section{Conclusion}\label{sec::conclusion}

In this paper, we study self-supervised deep graph clustering and propose a novel method termed R$^2$FGC. R$^2$FGC introduces the relational learning for the graph-structured data, in which the attribute- and structure-level relation information among nodes are extracted based on AE and GAE. To achieve effective representations, R$^2$FGC preserves consistent relations among the nodes under augmentation, whereas the redundancy relation is filtered for discriminative representations. R$^2$FGC also cooperates a representation fusion mechanism with the relational learning to instruct downstream self-supervised clustering tasks jointly. Experimental results on various benchmark datasets demonstrate the validity and superiority of the proposed method. In the future, we aim to extend relational learning to other scenarios including multi-view graph clustering, interpretable clustering, and other promising applications such as face clustering and text clustering.

\section*{Acknowledgments}
The authors are grateful to the anonymous reviewers for critically reading the manuscript and for giving important suggestions to improve their paper.



\bibliographystyle{IEEEtran}
\bibliography{ref.bib}

\begin{IEEEbiography}
[{\includegraphics[width=1in,height=1.25in]{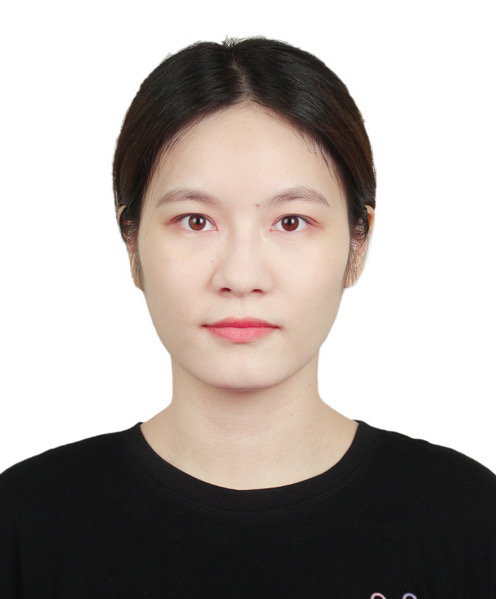}}]
{Siyu Yi} is currently a Ph.D. candidate in statistics from Nankai University, Tianjin, China. She received the B.S. and M.S. degrees in Mathematics from Sichuan University, Sichuan, China, in 2017 and 2020, respectively. Her research interests focus on graph representation learning, design of experiments, and subsampling in big data. 
\end{IEEEbiography}

\begin{IEEEbiography}
[{\includegraphics[width=1in,height=1.25in]{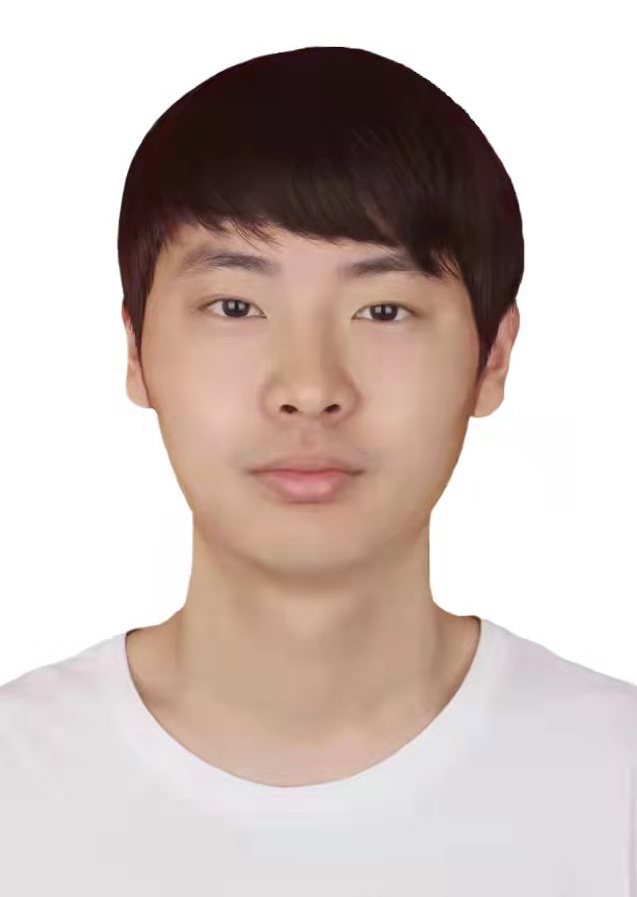}}]
{Wei Ju} is currently a postdoc research fellow in Computer Science at Peking University. Prior to that, he received his Ph.D. degree in Computer Science from Peking University, Beijing, China, in 2022. He received the B.S. degree in Mathematics from Sichuan University, Sichuan, China, in 2017. His current research interests lie primarily in the area of machine learning on graphs including graph representation learning and graph neural networks, and interdisciplinary applications such as knowledge graphs, drug discovery and recommender systems. He has published more than 30 papers in top-tier venues and has won the best paper finalist in IEEE ICDM 2022.
\end{IEEEbiography}

\begin{IEEEbiography}
[{\includegraphics[width=1in,height=1.25in]{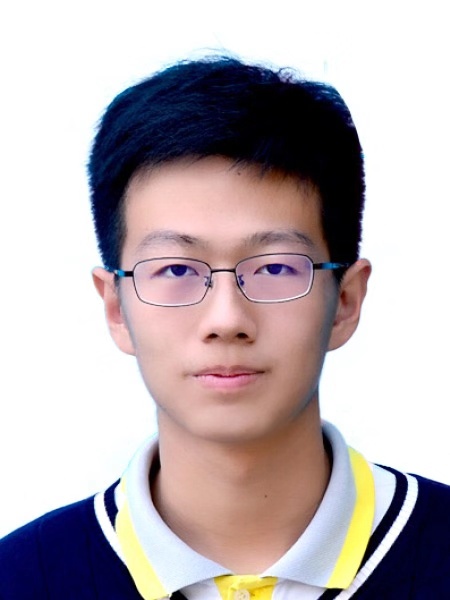}}]
{Yifang Qin} is an graduate student in School of Computer Science, Peking University, Beijing, China. Prior to that, he received the B.S. degree in school of EECS, Peking University. His research interests include graph representation learning and recommender systems.
\end{IEEEbiography}

\begin{IEEEbiography}
[{\includegraphics[width=1in,height=1.25in]{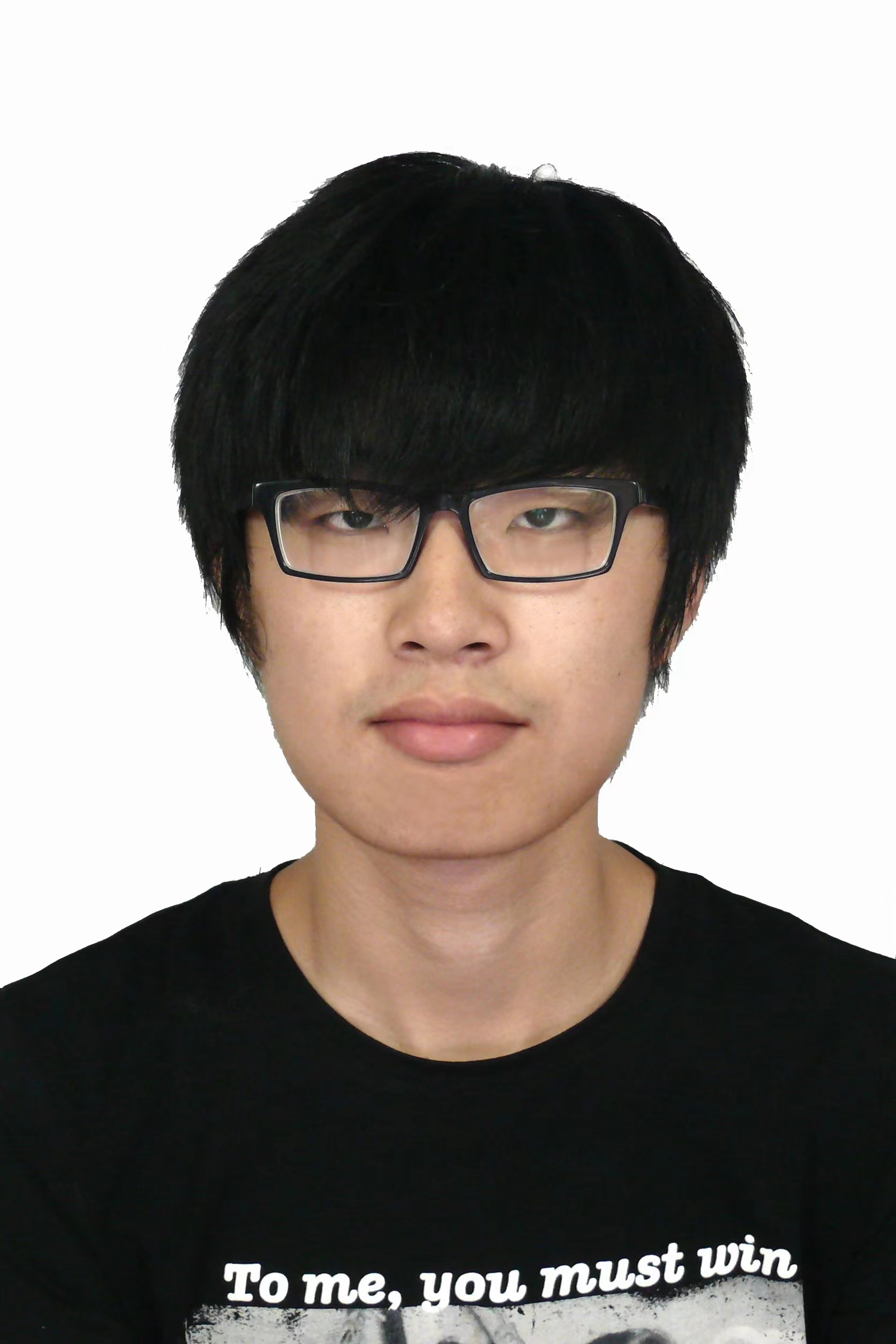}}]
{Xiao Luo} is a postdoctoral researcher in Department of Computer Science, University of California, Los Angeles, USA. Prior to that, he received the Ph.D. degree in School of Mathematical Sciences from Peking University, Beijing, China and the B.S. degree in Mathematics from Nanjing University, Nanjing, China, in 2017. 
His research interests includes machine learning on graphs, image retrieval, statistical models and bioinformatics. 
\end{IEEEbiography}

\begin{IEEEbiography}
[{\includegraphics[width=1in,height=1.25in]{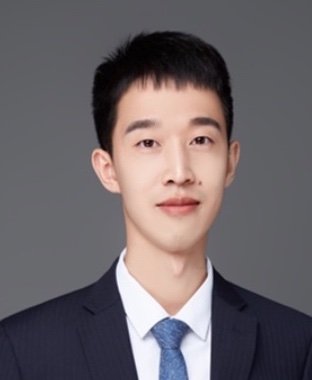}}]
{Luchen Liu} is currently a post-doctoral research fellow in Computer Science at Peking University. He received the Ph.D. degree in Computer Science from Peking University in 2020. His current research interests lie primarily in the area of deep learning for temporal graph data and interdisciplinary applications such as intelligent healthcare and quantitative investment.
\end{IEEEbiography}

\begin{IEEEbiography}
[{\includegraphics[width=1in,height=1.25in]{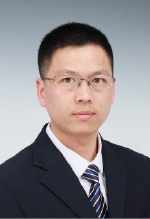}}] {Yongdao Zhou} received his B.S. degree in Mathematics, M.S. and Ph.D. degree
in Statistics from Sichuan University, China, in 2002, 2005, and 2008, respectively.  After graduation, he joined Sichuan University and was a professor after 2015. In 2017, he then joined Nankai University, where he is presently a professor in Statistics. His research agenda focuses on design of experiments and big data analysis. He published more than 60 papers and 5 monographs. His research publications have won best paper awards in WCE 2009 and Sci Sin Math in 2023. 
\end{IEEEbiography}

\begin{IEEEbiography}
[{\includegraphics[width=1in,height=1.25in]{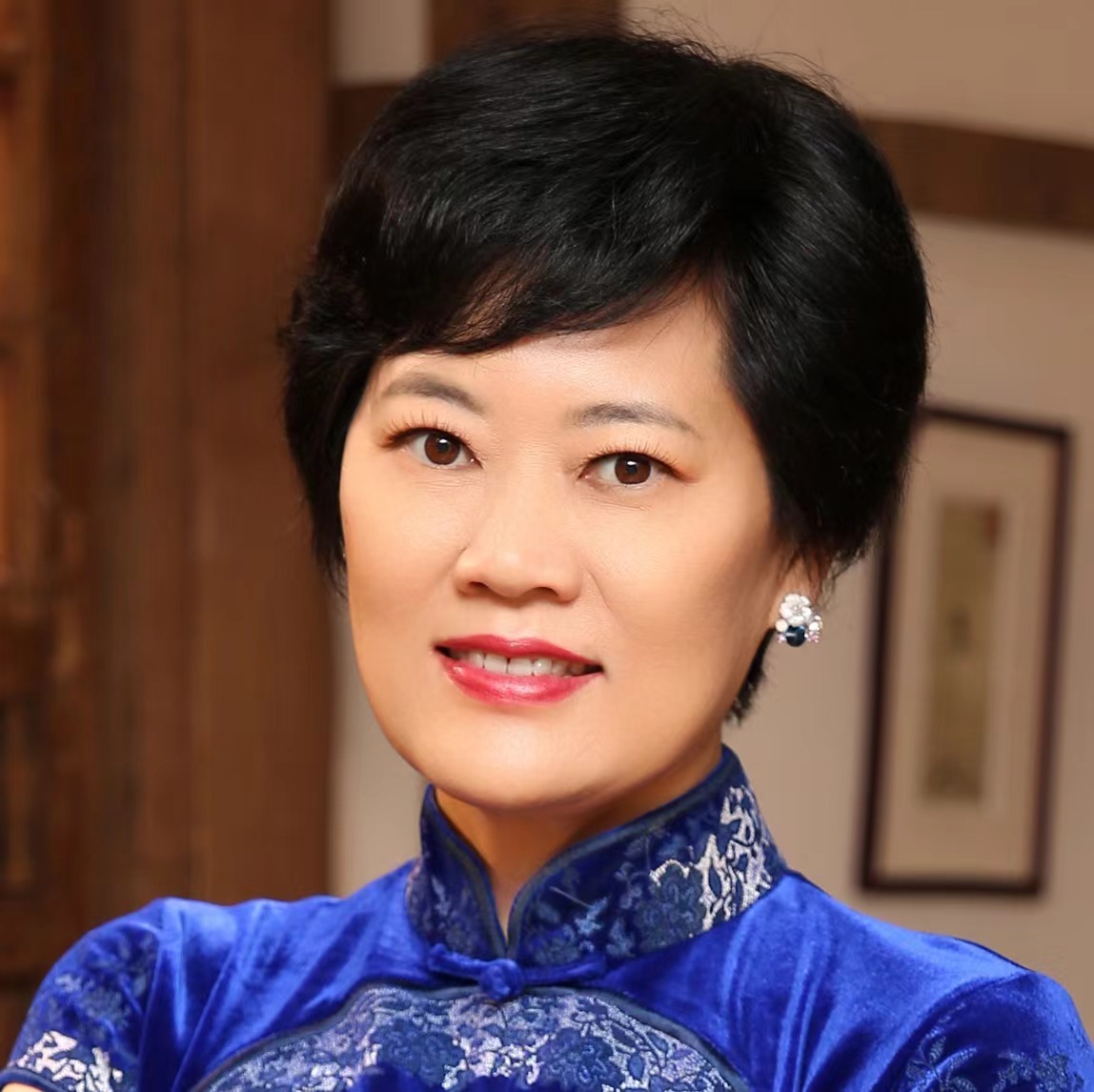}}]
{Ming Zhang} received her B.S., M.S. and Ph.D. degrees in Computer Science from Peking University respectively. She is a full professor at the School of Computer Science, Peking University. Prof. Zhang is a member of Advisory Committee of Ministry of Education in China and the Chair of ACM SIGCSE China. She is one of the fifteen members of ACM/IEEE CC2020 Steering Committee. She has published more than 200 research papers on Text Mining and Machine Learning in the top journals and conferences. She won the best paper of ICML 2014 and best paper nominee of WWW 2016. Prof. Zhang is the leading author of several textbooks on Data Structures and Algorithms in Chinese, and the corresponding course is awarded as the National Elaborate Course, National Boutique Resource Sharing Course, National Fine-designed Online Course, National First-Class Undergraduate Course by MOE China.
\end{IEEEbiography}

\vfill

\end{document}